\documentclass[lettersize,journal]{IEEEtran}
\usepackage{amsmath,amsfonts}
\usepackage{algorithmic}
\usepackage{array}
\usepackage[caption=false,font=normalsize,labelfont=sf,textfont=sf]{subfig}
\usepackage{textcomp}
\usepackage{stfloats}
\usepackage{url}
\usepackage{verbatim}
\usepackage{graphicx}
\def\BibTeX{{\rm B\kern-.05em{\sc i\kern-.025em b}\kern-.08em
    T\kern-.1667em\lower.7ex\hbox{E}\kern-.125emX}}
\usepackage{balance}
\usepackage{amssymb}
\usepackage{comment}
\usepackage{tcolorbox}
\usepackage{booktabs}
\usepackage{multicol}
\usepackage{todonotes}
\usepackage{multirow}
\usepackage{hyperref}

\DeclareFontFamily{U}{BOONDOX-calo}{\skewchar\font=45 }
\DeclareFontShape{U}{BOONDOX-calo}{m}{n}{
  <-> s*[1.05] BOONDOX-r-calo}{}
\DeclareFontShape{U}{BOONDOX-calo}{b}{n}{
  <-> s*[1.05] BOONDOX-b-calo}{}
\DeclareMathAlphabet{\mathcalboondox}{U}{BOONDOX-calo}{m}{n}
\SetMathAlphabet{\mathcalboondox}{bold}{U}{BOONDOX-calo}{b}{n}
\DeclareMathAlphabet{\mathbcalboondox}{U}{BOONDOX-calo}{b}{n}
\usepackage{relsize}

\begin{document}

\title{A signal processing interpretation of noise-reduction convolutional neural networks}

\author{Luis A. Zavala-Mondragón \IEEEmembership{Student member IEEE}, Peter H.N. de With, \IEEEmembership{Fellow IEEE}, \\and Fons van der Sommen, \IEEEmembership{Member IEEE}
	\thanks{Luis A. Zavala-Mondragon, Fons van der Sommen and P.H.N de With are with the Electrical Engineering department of the Eindhoven University of Technology, VCA lab, Flux 5, Eindhoven (corresponding e-mail: lzavala905@gmail.com).}
}

\markboth{Journal of \LaTeX\ Class Files,~Vol.~14, No.~8, August~2021}%
{Shell \MakeLowercase{\emph{et al.}}: A Sample Article Using IEEEtran.cls for IEEE Journals}

\IEEEpubid{0000--0000/00\$00.00~\copyright~2021 IEEE}

\maketitle

\begin{abstract}
Encoding-decoding CNNs play a central role in data-driven noise reduction and can be found within numerous deep-learning algorithms. However, the development of these CNN architectures is often done in \emph{ad-hoc} fashion and theoretical underpinnings for important design choices is generally lacking. Up to this moment there are different existing relevant works that strive to explain the internal operation of these CNNs. Still, these ideas are either scattered and/or may require significant expertise to be accessible for a bigger audience. In order to open up this exciting field, this article builds intuition on the theory of deep convolutional framelets and explains diverse ED CNN architectures in a unified theoretical framework. By connecting basic principles from signal processing to the field of deep learning, this self-contained material offers significant guidance for designing robust and efficient novel CNN architectures. 
\end{abstract}


%%%%%%%%%%%%%%%%%%%%%%%%%%%%%%%%
\section{Introduction}
%%%%%%%%%%%%%%%%%%%%%%%%%%%%%%%%
%
A well-known image-processing application is noise/artifact reduction of images, which is consists in estimating a noise/artifact-free signal out of a noisy observation. In order to achieve this, conventional signal processing algorithms often employ explicit assumptions on the signal and noise characteristics, which has resulted in well-known algorithms such as wavelet shrinkage~\cite{chang2000adaptive}, sparse dictionaries~\cite{elad2006image}, total-variation minimization~\cite{rudin1992nonlinear} and low-rank approximation~\cite{jin2015annihilating}. With the advent of deep learning techniques, signal processing algorithms applied to image denoising, have been regularly outperformed and increasingly replaced by encoding-decoding convolutional neural networks (CNNs).

In this article, rather than conventional signal processing algorithms, we focus on the so-called encoding-decoding CNNs. These models contain an \emph{encoder} which maps the input to a multi-channel/redundant representations and a \emph{decoder}, which maps the encoded signal back to the original domain. In both, the encoder and decoder, sparsifying non-linearities which suppress parts of the signal are applied. In contrast with conventional signal processing algorithms, encoding-decoding CNNs are often presented as a solution which does not make explicit assumptions on the signal and noise. For example, in supervised algorithms, an encoding-decoding CNN \emph{learns} the optimal parameters to filter the signal from a set of paired examples of noise/artifact-free images and images contaminated with noise/artifacts~\cite{chen2017low, han2018framing, Zhang2017}, which highly simplifies the solution of the noise-reduction problems, since this circumvents the use of explicit modeling of the signal and noise. Furthermore, the good performance and simple use of encoder-decoder CNNs/autoencoders have enabled additional data-driven noise-reduction algorithms, where CNNs are embedded as part of a larger system. Examples of such approaches are unsupervised noise-reduction~\cite{yokota2020manifold}, denoising based on generative adversarial networks~\cite{kusters2021conditional}. Besides this, smoothness in signals can be obtained also by advanced regularization using CNNs, e.g. by exploiting data-driven model-based iterative reconstruction~\cite{gupta2018cnn}.

Despite of the impressive noise-reduction performance and flexibility of encoding-decoding convolutional neural networks, these models have also downsides that should be considered. First, the complexity and heuristic nature of such designs often offers restricted understanding of the internal operation of such architectures~\cite{McCann2017convolutional}. Second, the  training and deployment of CNNs requires specialized hardware and the use of significant computational resources. Third and final, the restricted understanding of the signal modeling in encoding-decoding CNNs does not clearly reveal the limitations of such models and, consequently, it is not obvious how to overcome these problems.

In order to overcome the limitations of encoding-decoding CNNs, new research has tackled the lack of explainability of these models by acknowledging the similarity of the building blocks of encoding-decoding CNNs applied to image noise reduction and the elements of well-known signal processing algorithms, such as wavelet decomposition, low-rank approximation~\cite{ye2018deep, ye2019understanding, Mohan2020Robust}, variational methods~\cite{unser2022kernel}, lower-dimensional manifolds~\cite{yokota2020manifold} and convolutional sparse coding~\cite{papyan2017convolutional}. Furthermore, practical works based on shrinkage-based CNNs inspired in well-established wavelet shrinkage algorithms has further deepened the connections between signal processing and CNNs~\cite{mentl2017noise,Fan2020}. This unified treatment of signal processing-inspired CNNs has resulted in more explainable~\cite{han2018framing, yokota2020manifold}, better performing~\cite{han2018framing} and more memory-efficient designs~\cite{zavala2022noise}.

\IEEEpubidadjcol
This article has three main objectives. First, to summarize the diverse explanations of the components of encoding-decoding convolutional neural networks applied to image noise reduction based on the concept of deep convolutional framelets~\cite{ye2018deep} and on elementary signal processing concepts. Both aspects are considered with the aim of achieving an in-depth understanding of the internal operation of encoding-decoding CNNs and to show that the design choices have \emph{implicit} assumptions about the signal behavior inside the CNN. A second objective is to offer practitioners tools for optimizing their CNN designs with signal processing concepts. Third and final, the aim is to show practical use cases, where existing CNNs are analyzed in a unified framework, thereby enabling a better comparison of different designs by making their internal operation explicitly visible. Our analysis are based on existing works~\cite{ye2018deep, han2018framing, zavala2021image}, who analyzed CNNs where the non-linearities are ignored. In this article, we overcome this limitation and present a complete analysis including the non-linear activations, which reveals important assumptions implicit in the analyzed models.

The structure of this article is as follows. Section~\ref{sec:chapter4_notation} introduces the notation that is used in this text. Section~\ref{sec:chapter4_encodingDecoding} describes the signal model and the architecture of encoding-decoding networks. Afterwards, Section~\ref{sec:chapter4_fundamentalsOfSP} addresses fundamental aspects of signal processing, such as singular value decomposition, low-rank approximation, framelets, as well as the estimation of signals in the framelet domain. All the concepts of Sections~\ref{sec:chapter4_encodingDecoding} and \ref{sec:chapter4_fundamentalsOfSP} converge in Section~\ref{sec:chapter4_tdcfAndShrinkage}, where the encoding-decoding CNNs are interpreted in terms of a data-driven low-rank approximation and of wavelet shrinkage. Afterwards, based on the learnings from Section~\ref{sec:chapter4_tdcfAndShrinkage}, Section~\ref{sec:chapter4_analysis} shows the analysis of diverse architectures from a signal processing perspective and under a set of explicit assumptions. Afterwards, Section~\ref{sec:trainedModel} explores if some of the theoretical properties exposed here are related to trained models. Based on the diverse described models and  theoretical operation of CNNs, Section~\ref{sec:designElements} addresses a design criterion which can be used to design or choose new models and briefly describes the state-of-the art for noise reduction with CNNs. Finally, Section~\ref{sec:chapter4_conclussions} elaborates concluding remarks and discusses diverse elements that have not yet been (widely) explored by current CNN designs. 
%
%
%
%
%%%%%%%%%%%%%%%%%%%%%%%%%%%%%%%%
\section{\textbf{Notation} }~\label{sec:chapter4_notation}
%%%%%%%%%%%%%%%%%%%%%%%%%%%%%%%%
%
Convolutional neural networks (CNNs) are composed by basic elements, such as convolution, activation and down/up-sampling layers. In order to achieve better clarity in the explanations given in this paper, we define the mathematical notation to represent the basic operations of CNNs. Part of the definitions presented here are based on the work of Zavala~\emph{et al.}~\cite{zavala2022noise}.

In the following, a scalar is represented by a lower-case letter (e.g. $a$), while a vector is represented by an underlined lower-case letter (e.g. $\underline{a}$). Furthermore, a matrix, such as an image or convolution mask, is represented by a boldface lowercase letter (e.g. variables $\mathbf{x}$ and $\mathbf{y}$). Finally, a tensor is defined by a boldface uppercase letter. For example, the two arbitrary tensors $\mathbf{A}$ and $\mathbf{Q}$ are defined by
\begin{equation}
    \mathbf{A} =
    \begin{pmatrix}
        \mathbf{a}^0_0 &  \dots & \mathbf{a}^0_{ N_{\text{C}}-1}\\
        \vdots & \ddots &  \vdots\\ 
        \mathbf{a}_0^{ N_{\text{R}} -1} & \dots  &  \mathbf{a}^{ N_{\text{R}}-1}_{ N_{\text{C}}-1}\\
    \end{pmatrix}, 
    \mathbf{Q} =
    \begin{pmatrix}
        \mathbf{q}^0 \\ \vdots\\ \mathbf{q}^{ N_{\text{R}} -1}
    \end{pmatrix}.
\end{equation}
Here, entries $\mathbf{a}^r_c$ and $\mathbf{q}^r$ represent two-dimensional arrays (matrices). Since the defined tensors are used in the context of CNNs, matrices $\mathbf{a}^r_c$ and $\mathbf{q}^r$ are learned filters, which have dimensions $(N_{\text{V}} \times N_{\text{H}})$, where $N_{\text{V}}$ and $N_{\text{H}}$ denote the filter dimensions in the vertical and horizontal directions, respectively. Finally, we define the total tensor dimension of $\mathbf{A}$ and $\mathbf{Q}$ by $(N_{\text{C}} \times N_{\text{R}} \times N_\text{V} \times N_{\text{H}})$ and $(N_{\text{R}} \times 1 \times N_{\text{V}} \times N_{\text{H}})$, where $N_{\text{R}}$ and $ N_{\text{C}}$ are the number of row and column entries, respectively. If the tensor $\mathbf{A}$ contains the convolution weights in a CNN, the row-entry dimensions represent the input number of channels to a layer, while the number of column elements denotes the number of output channels.

Having defined the notation for the variables, we focus on a few relevant operators. First, the transpose of a tensor $(\cdot)^\intercal$, expressed by
\begin{equation}
    \mathbf{Q}^\intercal =
    \begin{pmatrix}
        \mathbf{q}^0 & \dots & \mathbf{q}^{N_{\text{R}} - 1}
    \end{pmatrix}.
\end{equation}
Furthermore, the convolution of two tensors is written as $\mathbf{A}\mathbf{Q}$ and specified by
\begin{equation}
    \mathbf{A} \mathbf{Q} = 
        \begin{pmatrix}
        \sum_{r=0}^{N_R-1} \mathbf{a}_r^0*\mathbf{q}^r \\
        \vdots\\
        \sum_{r=0}^{N_R-1}\mathbf{a}_r^{N_{\text{R}}-1}*\mathbf{q}^{r}
    \end{pmatrix}.
\end{equation}
Here, the symbol $*$ defines the convolution between two matrices (images).

In this paper, images which are 2D arrays (matrices), are often convolved with 4D tensors. When this operation is performed, images are considered to have dimensions $(1 \times 1 \times N_\text{V} \times N_\text{H})$. In addition, in this paper matrix $\mathbf{I}$ is the identity signal for the convolution operator, which for a 2D image is the Kronecker delta/discrete impulse (an image with a single non-zero pixel with unity amplitude at the center of the image). Furthermore, we indicate that variables in the \emph{decoding path} of a CNN are distinguished with a tilde (e.g. $\tilde{\mathbf{K}}$, $\underline{\tilde{b}}$).

Additional symbols that will be used throughout the article are the down- and up-sampling operations by a factor $s$, which are denoted by $f_{(s\downarrow)}(\cdot)$ for down-sampling and for up-sampling $f_{(s\uparrow)}(\cdot)$. In this paper, both operations are defined in the same way as in multi-rate filter banks. For example, consider the signal 
\begin{equation}
    \underline{x} =
    \begin{pmatrix}
         1,&  2,&  3, & 4, & 5, &6, &7, &8, & 9, & 10
    \end{pmatrix}.
\end{equation}
If we apply the down-sampling operator to $\underline{x}$ by a factor 2, this results into 
\begin{equation}
    \underline{z} = f_{(2\downarrow)}(\underline{x})= 
    \begin{pmatrix}
       1,&  3, & 5, &7, & 9
    \end{pmatrix},
\end{equation}
where $\underline{z}$ is the down-sampled version of $\underline{x}$. Conversely, the result of applying the up-sample operator $f_{(2\uparrow)}(\cdot)$ gives as result
\begin{equation}
    f_{(2\uparrow)}(\underline{z})=
    \begin{pmatrix}
        1,& 0,&  3, & 0, & 5, & 0, & 7, & 0, & 9, & 0
    \end{pmatrix}.
\end{equation}
Additional operators used in the article are the \emph{rectified linear unit} (ReLU), the \emph{shrinkage/thresholding} and the \emph{clipping}, which are represented by $(\cdot)_+$, $\tau_{(\cdot)}(\cdot)$ and $\mathcal{C}_{(\cdot)}(\cdot)$, respectively.

\begin{figure}
    \centering
    \includegraphics[width=0.5\columnwidth]{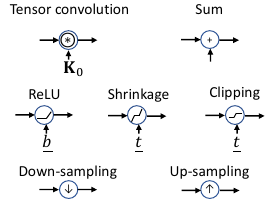}
    \caption{Symbols used for the schematic representations of the CNNs addressed in this article.}
    \label{fig:fig1}
\end{figure}

For better clarity, the most important symbols used in this article are summarized in Table~\ref{tab:symbols}. In addition, the graphical representations of some of the symbols that will be used to graphically describe CNNs are shown in Fig.~\ref{fig:fig1}.
\begin{table}[!h]
	\caption{Relevant symbols used in this paper}\label{tab:symbols}
	\centering
	\begin{tabular}{ c | p{0.75\columnwidth} }
		\toprule
		Symbol & Meaning \\ \midrule
		$f_{(2\downarrow)}(\cdot)$   & Down-sampling operation.\\
		$f_{(2\uparrow)}(\cdot)$   & Up-sampling operation.\\
		$\mathbf{I}$ & Convolution identity.\\
        $\mathbf{K}$ & Encoding convolution kernel. \\
		$\tilde{\mathbf{K}}$ & Decoding convolution kernel. \\
		$\mathbf{W}$   &Filters for the forward discrete wavelet transform. \\
		$\tilde{\mathbf{W}}$  &Filters for the inverse discrete wavelet transform. \\
		$\mathbf{W}_\text{H}$  &High-pass filters of the forward discrete wavelet transform. \\
		$\mathbf{W}_\text{L}$  &Low-pass filter of the forward discrete wavelet transform. \\
		$\mathbf{x}$ & Noiseless image.\\
		$\mathbf{y}$ & Noisy image.\\
		$\boldsymbol\eta$ & Additive noise. \\
		$\underline{b}$ & Bias vector.\\
        $t$ & Threshold level.\\
		$*$ & Image convolution.\\
		$\mathbf{K}\mathbf{x}$ & Tensor convolution between tensor $\mathbf{K}$ and signal $\mathbf{x}$. \\
		$(\cdot)^\intercal$ & Transpose of a tensor. \\
		$(\cdot)_{+}$ &  ReLU activation.\\
		$\tau_{(\cdot)}(\cdot)$ & Generic thresholding/shrinkage operation.\\
		$\mathcal{C}_{(\cdot)}(\cdot)$ & Generic clipping operation.\\
		\bottomrule
	\end{tabular}
\end{table}
%
%%%%%%%%%%%%%%%%%%%%%%%%%%%%%%%%
\section{\textbf{Encoding-decoding CNNs} }~\label{sec:chapter4_encodingDecoding}
%%%%%%%%%%%%%%%%%%%%%%%%%%%%%%%%
%
%%%%%%%%%%%%%%%%%%%%%%%%%%%%%%%%
\subsection{\textbf{Signal model and noise reduction configurations} }~\label{sec:chapter4residualcnns} %%
\begin{figure}[h!]
    \centering
    \includegraphics[width=0.8\columnwidth]{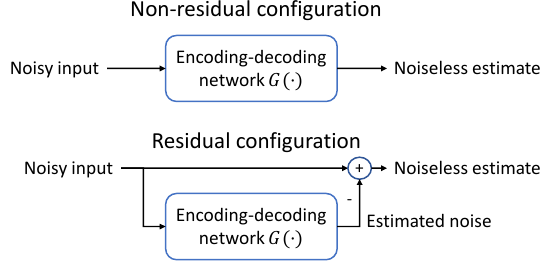}
    \caption{Residual and non-residual network configurations. Note that the main difference between both designs is the global skip connection occurring in the residual structure. Still, it can be observed that the network $G(\cdot)$ may contain skip connections \emph{internally}.}
    \label{fig:fig2}
\end{figure}

%%%%%%%%%%%%%%%%%%%%%%%%%%%%%%%%
%
In noise-reduction applications, the common additive signal model is defined by
\begin{equation}
    \mathbf{y} = \mathbf{x} + \boldsymbol\eta,
\end{equation}
where the observed signal $\mathbf{y}$ is the result of contaminating a noiseless image $\mathbf{x}$ with additive noise $\boldsymbol\eta$. Assume that the noiseless signal $\mathbf{x}$ is to be estimated from the noisy observation $\mathbf{y}$. In deep learning applications, this is often achieved by models with the form
\begin{equation}
    \hat{\mathbf{x}} = G(\mathbf{y}).
    \label{eq:chapter4NonResidualLearning}
\end{equation}
Here, $G(\cdot)$ is a generic encoding-decoding CNN. We refer to this form of noise reduction as \emph{non-residual}. Alternatively, it is possible to find $\hat{\mathbf{x}}$ by training $G(\cdot)$ to estimate the noise component $\hat{\boldsymbol\eta}$, and subtract it from the noisy image $\mathbf{y}$ to estimate the noiseless image $\hat{\mathbf{x}}$, or equivalently
\begin{equation}
    \hat{\mathbf{x}} = \mathbf{y} - G(\mathbf{y}).
    \label{eq:chapter4residualLearning}
\end{equation}
This model is referred to as \emph{residual}~\cite{chen2017low,jin2017deep,Zhang2017}, because the output of the network is subtracted from its input. For reference, Fig.~\ref{fig:fig2} portrays the difference of the placement of the encoding-decoding structure in residual and non-residual configurations. 
%
%
%%%%%%%%%%%%%%%%%%%%%%%%%%%%%%%%%%%%%%%%
\subsection{\textbf{Encoding-decoding CNNs} }
%%%%%%%%%%%%%%%%%%%%%%%%%%%%%%%%%%%%%%%%
%
%
Encoding-decoding (convolutional) neural networks are rooted in the techniques for data-dimensionality reduction and unsupervised feature extraction, where a given signal is mapped to an alternative space via a non-linear transformation. This space should have properties which are somehow attractive for the considered task. For example, for dimensionality reduction, the alternative space should be lower-dimensional than the original input. In this article, we are interested in models that are useful for noise-reduction applications. Specifically,  this manuscript addresses models that are referred to as encoding-decoding \emph{convolutional} neural networks such as the model by Ranzato~\emph{et al.}~\cite{ranzato2007unsupervised}, in which the encoder uses convolution filters to produce multi-channel/redundant representations, in which sparsifying non-linearities are applied. The sparsified signal is later mapped back to the original representation. It should be noted that despite the fact that the origins of the encoding-decoding CNNs are linked to feature extraction, this type of architecture quickly showed to be useful for other applications such as noise reduction, which is the topic of this article. For the rest of this manuscript whenever we mention an encoding-decoding CNN, we are referring to a design which follows the same basic principles of Ranzato's design. 

It can be observed that encoding-decoding CNNs are constituted of three main parts. (1) The \textbf{encoder}, which maps the incoming image to a representation with more image channels with a convolution layer. Every channel of the resulting redundant representation contains a fraction of the content of the original signal. It should be noted that the encoder often (but not necessarily) decreases the resolution of the higher dimensional representation, to enable multi-resolution processing and to decrease the memory requirements of the design. (2) The \textbf{decoder}, which maps the multi-channel representation back to the original space. (3) The \textbf{non-linearities}, which suppress specific parts of the signal. In summary, the most basic encoding-decoding step in a CNN $G(\cdot)$ is expressed by
\begin{equation}
    \boxed{
    	G(\mathbf{y})= G_\text{dec}(
    	    G_\text{enc}(\mathbf{y})
    	)
    }\ \ ,
    \label{eq:chapter4encodingDecoding}
\end{equation}
where $G_\text{enc}(\cdot)$ is the encoder, which is generally defined by
\begin{equation}
    \begin{aligned}
        \mathbf{C}_0 = &E_0(\mathbf{y}),\\
        \mathbf{C}_1 = &E_1(\mathbf{\mathbf{C}}_0),\\
        \mathbf{C}_2 = &E_2(\mathbf{\mathbf{C}}_1),\\
            \vdots \\
        \mathbf{C}_{N-1} = &E_{N-1}(\mathbf{\mathbf{C}}_{N-2}),\\
     G_\text{enc}(\mathbf{y}) = &\mathbf{C}_{N-1}. \\
    \end{aligned}
\end{equation}
Here, $\mathbf{\mathbf{C}}_n$ represents the code generated by the $n$-th encoding $E_n(\cdot)$, which can be expressed by the equation
\begin{equation}
    \boxed{
    \mathbf{C}_{n} =
    E_n(\mathbf{C}_{n-1}) =  f_{(s\downarrow)}
    \big(
        A_{(\underline{b}_{n-1})}(
            \mathbf{K}_{n-1}
            \mathbf{C}_{n-1}
        )
    \big)
    }\ \ .
    \label{eq:chapter4encoder}
\end{equation}
Here, the function $A(\cdot)_{(\cdot)}$ is a generic activation used in the encoder and $f_{(s\downarrow)}(\cdot)$ is a down-sampling function by factor $s$. Complementary to the encoder, the decoder network maps the multi-channel sparse signal back to the original domain. Here, we define  the decoder by
\begin{equation}
    \begin{aligned}
    \tilde{\mathbf{\mathbf{C}}}_{N-2} = &D_{N-1}(\mathbf{\mathbf{C}}_{N-1}),\\
    \vdots \\
    \tilde{\mathbf{\mathbf{C}}}_1 = &D_2( \tilde{\mathbf{\mathbf{C}}}_2),\\
    \tilde{\mathbf{\mathbf{C}}}_0 = &D_1( \tilde{\mathbf{\mathbf{C}}}_1),\\
    G(\mathbf{y})= &D_0( \tilde{\mathbf{\mathbf{C}}}_0), 
    \end{aligned}
\end{equation}
where $\hat{\mathbf{\mathbf{C}}}_{n}$ is the $n-$th decoded signal, which is produced by the $n$-th decoder layer, yielding the general expression:
\begin{equation}
    \boxed{
    	\tilde{\mathbf{C}}_{n-1}
         =
         D_n(\tilde{\mathbf{C}}_{n})=
    	\tilde{A}_{(\tilde{\underline{b}})}
        \big(
            \tilde{\mathbf{K}}^\intercal_n
    	    f_{(s\uparrow)}(
    	        \tilde{\mathbf{C}}_{n}
            )
        \big)
    }\ \ .
    \label{eq:chaper4decoder}
\end{equation}
In the above, $\tilde{A}(\cdot)_{(\cdot)}$ is the activation function used in the decoder and $f_{(s\uparrow)}(\cdot)$ is an up-sampling function of factor $s$.

An important remark is that the encoder-decoder CNN does not always contain  down/up-sampling layers in which case, the decimation factor $s$ is unity, which  causes $f_{(1\uparrow)}(\mathbf{x}) = f_{(1\downarrow)}(\mathbf{x})=\mathbf{x}$ for any matrix $\mathbf{x}$. Furthermore, it should be noted also that we assume that the number of channels of the code $\mathbf{C}_{N}$ is always larger than the previous one $\mathbf{C}_{N-1}$. Furthermore,  it should be noted that a single encoder layer $E_n(\cdot)$ and its corresponding decoder layer $D_n(\cdot)$ can be considered a single-layer encoder-decoder network/pair.

For this article, the encoding convolution filter for a given layer $\mathbf{K}$ has dimensions $(N_\text{o} \times N_\text{i}\times N_\text{h} \times N_\text{v})$, where $N_\text{i}$  and $N_\text{o}$ are the number of input and output channels for a convolution layer, respectively. Similarly, $N_\text{h}$ and $N_\text{v}$ are the number of elements in the horizontal and vertical directions, respectively. Note that the encoder increases the number of channels of the signal (e.g. $N_\text{o} > N_\text{i}$), akin to Ranzatto's design~\cite{ranzato2007unsupervised}. Furthermore, it is assumed that the \emph{decoder} is symmetric in the number of channels to the \emph{encoder}, therefore, the dimensions of the decoding convolution kernel $\tilde{\mathbf{K}}^\intercal$ are $(N_\text{i} \times N_\text{o}\times N_\text{h} \times N_\text{v})$. The motivation of this symmetry is to emphasize the similarity between the signal processing and the CNN elements.
%
%
%
%
%%%%%%%%%%%%%%%%%%%%%%%%%%%%%%%%
\section{\textbf{Signal processing fundamentals} }~\label{sec:chapter4_fundamentalsOfSP}
%%%%%%%%%%%%%%%%%%%%%%%%%%%%%%%%
%
As shown by Ye~\emph{et al.}~\cite{ye2018deep}, within encoding-decoding CNNs, the signal is treated akin to well-known sparse representations, where the coefficients used for the transformation are directly learned from the training data. Prior to addressing this important concept in more detail, relevant supporting concepts such as sparsity, sparse transformations and non-linear signal estimation in the wavelet domain are explained.
%
%
%%%%%%%%%%%%%%%%
\subsection{\textbf{Sparsity} }
%%%%%%%%%%%%%%%%
%
A sparse image is a signal where most coefficients are small and the relatively few large coefficients capture most of the information~\cite{candes2008introduction}. This characteristic allows to discard low-amplitude components with relatively small perceptual changes. Hereby, the use of sparse signals is attractive for applications such as image compression, denoising and suppression of artifacts.

Despite the convenient characteristics of sparse signals, natural images are often non-sparse. Still, there are numerous transformations, which allow to map the signal to a sparse domain and which are analogous to the internal operations of CNNs. For example, \emph{singular value decomposition} factorizes the image in terms of two sets of orthogonal bases of which few basis pairs contain most of the energy of the image. An alternative transformation is based on \emph{framelets}, where an image is decomposed in a multi-channel representation, whereby each resulting channel contains a fragment of the Fourier spectrum. In the remainder of this section we will address all of these representations in more detail.
%
%
%%%%%%%%%%%%%%%%%%%%%%%%%%%%%%%%
\subsection{\textbf{Sparse signal representations} }
%%%%%%%%%%%%%%%%%%%%%%%%%%%%%%%%
%
%%%%%%%%%%%%%%%%%%%%%%%%%%%%%%%%
\subsubsection{\textbf{Singular value decomposition (SVD) and low-rank approximation}}\label{sec:chapter4SVD}
%%%%%%%%%%%%%%%%%%%%%%%%%%%%%%%%
%
Assume that an image (patch) is represented by a matrix $\mathbf{y}$ with dimensions $(N_\text{r} \times N_\text{c})$, where $N_\text{r}$ and $N_\text{c}$ are the number of rows and columns, respectively. Then, the singular value decomposition factorizes $\mathbf{y}$ as
\begin{equation}
    \mathbf{y} = \sum_{n=0}^{N_\text{SV}-1}  
    (\underline{u}_n \underline{v}_n^\intercal)
    \cdot
    \underline{\sigma}[n]
    ,
    \label{eq:chapter4svdVector}    
\end{equation}
in which $N_\text{SV}$ is the number of singular values, $n$ is a scalar index, while $\underline{u}_n$ and $\underline{v}_n$ are the $n^\text{th}$ left- and right-singular vectors, respectively. Furthermore, vector $\underline{\sigma}$ contains the singular values and each of its entries $\underline{\sigma}[n]$ is the weight assigned to every basis pair $\underline{u}_n$, $\underline{v}_n$. This means that the product $(\underline{u}_n\underline{v}_n^\intercal)$ contributes more to the image content for higher values of $\underline{\sigma}[n]$. It is customary that the singular values are ranked in descending order and the amplitudes of the singular values $\underline{\sigma}$ are sparse, therefore $\underline{\sigma}[0] \gg \underline{\sigma}[N_\text{SV}-1] $. The reason for this sparsity is because image (patches) intrinsically have high correlation. For example, many images contain repetitive patterns (e.g. a wall with bricks, a fence, the tiles of a rooftop or the stripes of a zebra) or uniform regions (for example, the sky, the skin of a person). This means an image patch may contain only a few linearly independent vectors that describe most of the image content. Consequently, a higher weight is assigned to such image bases.

Given that the amplitudes of the singular values of $\mathbf{y}$ in SVD are sparse, it is possible approximate $\hat{\mathbf{y}}$ with only a few bases $(\underline{u}_n\underline{v}_n^\intercal)$. Note that this procedure reduces the rank of signal $\mathbf{y}$ and hence it is known as \emph{low-rank approximation}. This process is equivalent to
\begin{equation}
    \hat{\mathbf{y}} = \sum_{n=0}^{N_\text{LR}-1}
    (
        \underline{u}_n 
        \underline{v}_n^\intercal
    )
    \cdot
    \sigma[n]
    ,
    \label{eq:chapter4lowRankApprox}    
\end{equation}
where $N_\text{SV} > N_\text{LR}$. Note that this effectively cancels the product $(\underline{u}_n  \underline{v}_n^\intercal)$ where the weight given by $\sigma[n]$ is low. Alternatively, it is possible to assign a weight of zero to the product $(\underline{u}_n  \underline{v}_n^\intercal)$ for $n \geq N_\text{LR}$.

\begin{figure}[!h]
    \centering
    \includegraphics[scale=0.8]{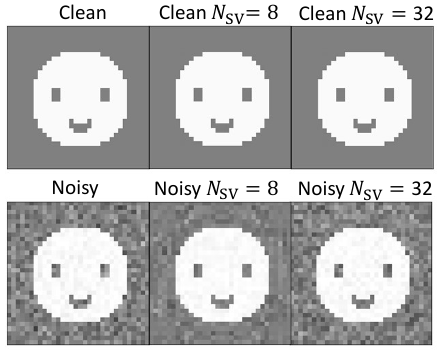}
    \caption{SVD reconstruction of clean and corrupted images with a different number of singular values. Note that the reconstruction of the clean image with 8 or 32 singular values ($N_\text{SV}=8$ or $N_\text{SV}=32$, respectively) yields to reconstructions indistinguishable from the original image. This contrasts with their noisy counterparts, where $N_\text{SV}=8$ reconstructs a smoother image in which the noise is attenuated, while $N_\text{SV}=32$ reconstructs the noise texture perfectly.
    }
    \label{fig:fig3}
\end{figure}

The low-rank representation of a matrix is desirable for diverse applications among which we can find image denoising. The motivation for using low-rank approximation for this application results from the fact that --as mentioned earlier-- natural images are considered low-rank due to the strong spatial correlation between pixels, whereas noise is high-rank (it is spatially uncorrelated). In consequence, reducing the rank/number of singular values decreases the presence of noise, while still providing a good approximation of the noise-free signal, as exemplified in Fig.~\ref{fig:fig3}. 
%
%
%%%%%%%%%%%%%%%%%%%%%%%%%%%%%%%%
\subsubsection{\textbf{Framelets}}\label{sec:chapter4framelets}
%%%%%%%%%%%%%%%%%%%%%%%%%%%%%%%%
%
Just as SVD, framelets are also commonly used for image processing.  In a nutshell, a framelet transform is a signal representation that factorizes/decomposes an arbitrary signal into multiple bands/channels. Each of these channels contain a segment of the energy of the original signal. In image and signal processing, the framelet bands are the result of convolving the analyzed signal with a group of discrete filters that have finite length/support. In this article, the most important characteristic that the filters of the framelet transform should comply with, is that the bands they generate capture \emph{all} the energy contained on the input to the decomposition. This is important to avoid the loss of information of the decomposed signal. In this text, we refer to framelets that comply with the previous characteristics as \emph{tight} framelets and the following paragraphs will describe this property in more detail.

In its decimated version, the framelet decomposition for \emph{tight} frames is represented by
\begin{equation}
    \mathbf{Y}_\text{fram} =
        f_{(2\downarrow)}(
            \mathbf{F}
            \mathbf{y}
        ),
        \label{eq:chapter4forFram}
\end{equation}
in which $\mathbf{Y}_\text{fram}$ is the decomposed signal and $\mathbf{F}$ is the framelet basis (tensor). Note that the signal  $\mathbf{Y}_\text{fram}$ has more channels than $\mathbf{y}$. Furthermore, the original signal $\mathbf{y}$ is recovered from $\mathbf{Y}_\text{fram}$ by
\begin{equation}
    \mathbf{y} = 
    \tilde{ \mathbf{F} }^\intercal
    f_{(2\uparrow)}(
        \mathbf{Y}_\text{fram}
    )
    \cdot c. 
    \label{eq:chapter4invFram}
\end{equation}
\begin{figure*}
    \centering
    \includegraphics[width=\textwidth]{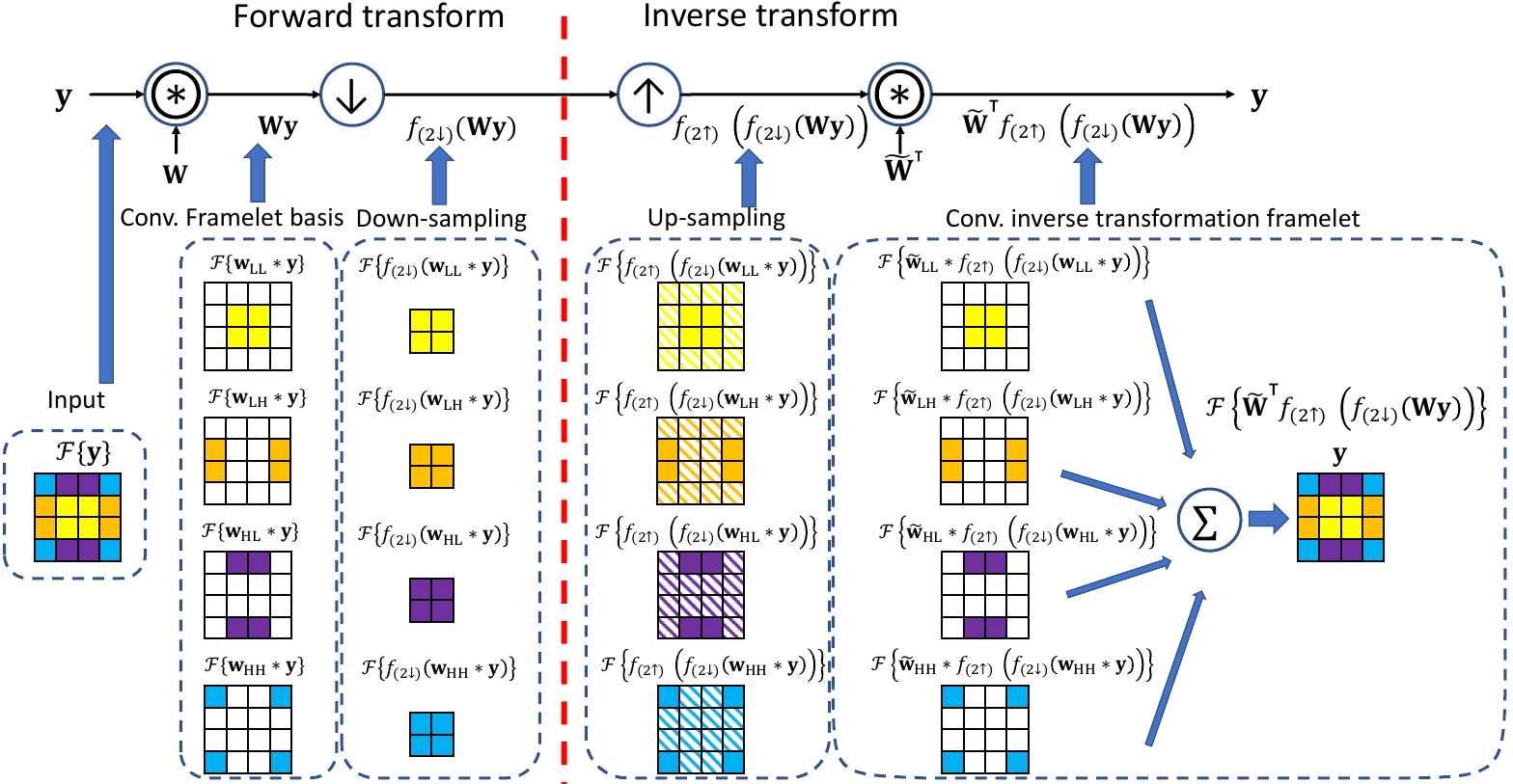}
    \caption{2D spectrum analysis of the decimated discrete framelet decomposition and reconstruction. In the figure, function $\mathcal{F}\{\cdot\}$ stands for the amplitude Fourier spectrum of the input argument. The yellow squares indicate a region in the low-frequency area of the Fourier spectrum, while the orange, purple and blue squares indicate the high-pass/detail bands. For these images, ideal orthogonal bases are assumed. Note that the forward transform is composed by two steps. First, the signal is convolved with the wavelet basis ($\mathbf{W}\mathbf{y}$). Afterwards, down-sampling is applied to the signal ($f_{(2\downarrow)}(\mathbf{W}\mathbf{y})$). During the inverse transformation, the signal is up-sampled by inserting zeros between each sample ($f_{(2\uparrow)}(f_{(2\downarrow)}(\mathbf{W}\mathbf{y}))$), which causes spatial aliasing (dashed blocks). Finally, the spatial aliasing is removed by the inverse transform filter $\tilde{\mathbf{W}}$ and all the channels are added ($\tilde{\mathbf{W}}^\intercal f_{(2\uparrow)}(f_{(2\downarrow)}(\mathbf{W}\mathbf{y}))$).
    }
    \label{fig:fig4}
\end{figure*}
Here, $\tilde{\mathbf{F}}$ is the filter of the inverse framelet transform and $c$ denotes an arbitrary constant. If $c=1$ the framelet is \emph{normalized}. Finally, note that the framelet transform can also be \emph{undecimated}. This means that in undecimated representations, the down-sampling and up-sampling layers $f_{(2\downarrow)} ( \cdot )$ and $f_{(2\uparrow)}(\cdot)$ are not used. An important property of the undecimated representation is that it is less prone to aliasing than its decimated counterpart, but more computationally expensive. Therefore, for efficiency reasons, the decimated framelet decomposition is often preferred over the undecimated representation. In summary, the decomposition and synthesis of the decimated framelet decomposition is represented by
\begin{equation}
    \boxed{
        \mathbf{y} = 
        \tilde{\mathbf{F}}^\intercal
        f_{(2\uparrow)}
        \big(
            f_{(2\downarrow)}
            \big(
                \mathbf{F}
                \mathbf{y}
            \big)
        \big)
        \cdot
        c
    }\ \ ,
    \label{eq:decimatedFramelet}
\end{equation}
while for the undecimated framelet it holds that
\begin{equation}
    \boxed{
        \mathbf{y} = 
        \tilde{\mathbf{F}}^\intercal
        (
            \mathbf{F}
            \mathbf{y}
        )
        \cdot
        c
    }\ \ .
    \label{eq:undecimatedFramelet}
\end{equation}

A notable normalized framelet is the \emph{discrete wavelet transform} (DWT), where variables $\mathbf{F}$ and $\tilde{\mathbf{F}}$ are replaced by tensors $\mathbf{W}=\begin{pmatrix}\mathbf{w}_\text{LL}, \mathbf{w}_\text{LH}, \mathbf{w}_\text{HL} , \mathbf{w}_\text{HH} \end{pmatrix}$ and 
$\tilde{\mathbf{W}} = \begin{pmatrix} \tilde{\mathbf{w}}_\text{LL}, \tilde{\mathbf{w}}_\text{LH}, \tilde{\mathbf{w}}_\text{HL}, \tilde{\mathbf{w}}_\text{HH} \end{pmatrix}$, respectively. Here, $\mathbf{w}_\text{LL}$ is the filter for the low-frequency band, while $\mathbf{w}_\text{LH}$, $\mathbf{w}_\text{HL}$, $\mathbf{w}_\text{HH}$ are the filters used to extract the detail in the horizontal, vertical and diagonal directions, respectively. Finally, $\tilde{\mathbf{w}}_\text{LH}$ $\tilde{\mathbf{w}}_\text{LH}$, $\tilde{\mathbf{w}}_\text{HL}$, $\tilde{\mathbf{w}}_\text{HH}$ are the filters of the inverse decimated DWT.

In order to understand the DWT more intuitively,  Fig.~\ref{fig:fig4} shows the decimated framelet decomposition using the filters of the discrete wavelet transform. Note that the convolution $\mathbf{W}\mathbf{y}$ results in a four-channel signal, where each channel contains only a fraction of the spectrum of image $\mathbf{y}$. This allows to down-sample each channel with minimal aliasing. Furthermore, to recover the original signal, each individual channel is up-sampled, thereby introducing aliasing, which is then removed by the filters of the inverse transform. Finally, all the channels are added and the original signal is recovered.

Analogous to the low-rank approximation, in framelets, the reduction of noise is achieved by setting noisy components to zero. These components are typically assumed to have low-amplitude compared with the amplitude of the sparse signal, as expressed by
\begin{equation}
    \hat{\mathbf{y}} = 
        \tilde{ \mathbf{F} }^\intercal
        f_{(2\uparrow)}
        \big(
            \tau_{ (\underline{t}) }
            \big(
                f_{(2\downarrow)} (
                    \mathbf{F}
                    \mathbf{y}
                )
            \big)
        \big)
        \cdot
        c
        ,
    \label{eq:chapter4SimpleShrinkage}
\end{equation}
where $\tau_{\underline{t} }(\cdot)$ is a generic thresholding/shrinkage function, which sets each of the pixels in $f_{(2\downarrow)}(\mathbf{F} \mathbf{y})$ to zero when values are lower than the threshold level $\underline{t}$.
%
%
%%%%%%%%%%%%%%%%%%%%%%%%%%%%%%%%
\subsection{\textbf{Nonlinear signal estimation in the framelet domain} }\label{sec:chapter4thresh}
%%%%%%%%%%%%%%%%%%%%%%%%%%%%%%%%
%
As mentioned in Section~\ref{sec:chapter4framelets}, framelets decompose a given image $\mathbf{y}$ by convolving it with a tensor $\mathbf{F}$. Note that many of the filters that compose $\mathbf{F}$ have a high-pass nature. Images often contain approximately uniform regions in which the variation is low, therefore, convolving a signal $\mathbf{y}$ with a high-pass filter $\mathbf{f}_\text{h}$ --where $\mathbf{f}_\text{h} \in \mathbf{F}$-- produces the sparse detail band $\mathbf{d}= \mathbf{f}_\text{h}*\mathbf{y}$ in which uniform regions have low amplitudes, while transitions i.e. edges contain most of the energy of the bands.

\begin{figure}[h!]
    \centering
    \includegraphics[width=\columnwidth]{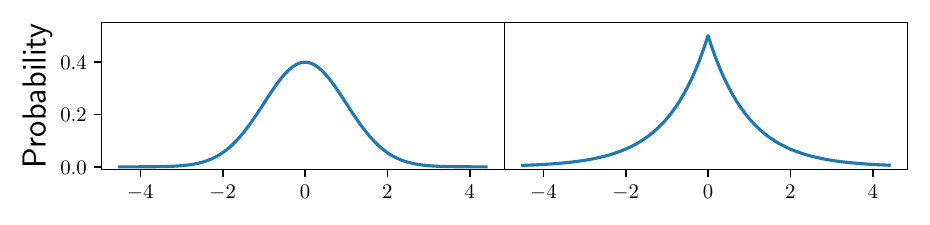}
    \caption{Probability density function for Gaussian (left) and Laplacian (right) distributions.}
    \label{fig:fig5}
\end{figure}

Assuming a model in which a single pixel $d\in\mathbf{d}$ is observed, which is contaminated with additive noise $\eta$. Then, the resulting observed pixel ${z}$ is defined by
\begin{equation}
	{z} = {d} + \eta.
	\label{eq:chapter4_noiseModelwaveletDomain}
\end{equation}
In order to recover the noiseless pixel ${d}$ from observation ${z}$, it is possible to use the point-\emph{maximum a posteriori} (MAP) estimate~\cite{chang2000adaptive, sendur2002bivariate}, defined by the maximization problem
\begin{equation}
    \hat { {d} } = \underset{ {d} }{\arg\text{max}} \big[ \ln \big(P( {d}| {z}) \big) \big].
    \label{eq:chapter4logMAP}
\end{equation}
Here, the log-posterior $\ln \big(P( {d}| {z}) \big)$ is defined by
\begin{equation}
	\ln \big( P( {d}| {z}) \big)  =
	\ln \big( P( {z}|{d}) \big)
	+
	\ln \big( P({d}) \big)
	\label{eq:chapter4logPosterior}
	,
\end{equation}
where the conditional probability density function (PDF) $ P({z}|{d})$ expresses the noise distribution, which is often assumed Gaussian and is defined by
\begin{equation}
    P({z}|{d}) \propto \exp \bigg(
        {-\frac{({z} - {d})^2}{2\sigma_{\eta}^2}}
    \bigg).
    \label{eq:chapter4gaussian}
\end{equation}
Here, $\sigma_{\eta}^2$ is the noise variance. Furthermore, as prior probability, it is assumed that the distribution of $P( {d})$ corresponds to a Laplacian distribution, which has been used in wavelet-based denoising~\cite{chang2000adaptive}. Therefore, $P( {d})$ is mathematically described by
\begin{equation}
	P({d}) \propto  \exp \bigg(
		-\frac{ |{d}| }{ \sigma_{d} }
	\bigg),
	\label{eq:chapter4prior}
\end{equation}
where $\sigma_{d}$ is the dispersion measure of the Laplace distribution. For reference, Fig.~\ref{fig:fig5} portrays an example of a Gaussian and a Laplacian PDF. Note that the Laplacian distribution has a higher probability of zero elements to occur than the Gaussian distribution for the same standard deviation.
Finally, substituting Eq.~(\ref{eq:chapter4gaussian}) and Eq.~(\ref{eq:chapter4prior}) in Eq~(\ref{eq:chapter4logPosterior}) results in
\begin{equation}
     \ln \big(P( {d} | {z}) \big) \propto -\frac{({z} - {d})^2}{2\sigma_{\eta}^2} -\frac{ | {d} | }{ \sigma_{d} }.
     \label{eq:chapter4_logProbabSolve}
\end{equation}
In the above, maximizing ${d}$ in $\ln \big(P( {d} | {z}) \big)$ with the first derivative criterion --in an (un)~constrained way--  leads to two common activations in noise-reduction CNNs: the ReLU and the soft-shrinkage function. Furthermore, the solution also can be used to derive the so-called \emph{clipping} function, which is useful in residual networks.

\begin{figure}[h!]
    \centering
    \includegraphics[width=\columnwidth]{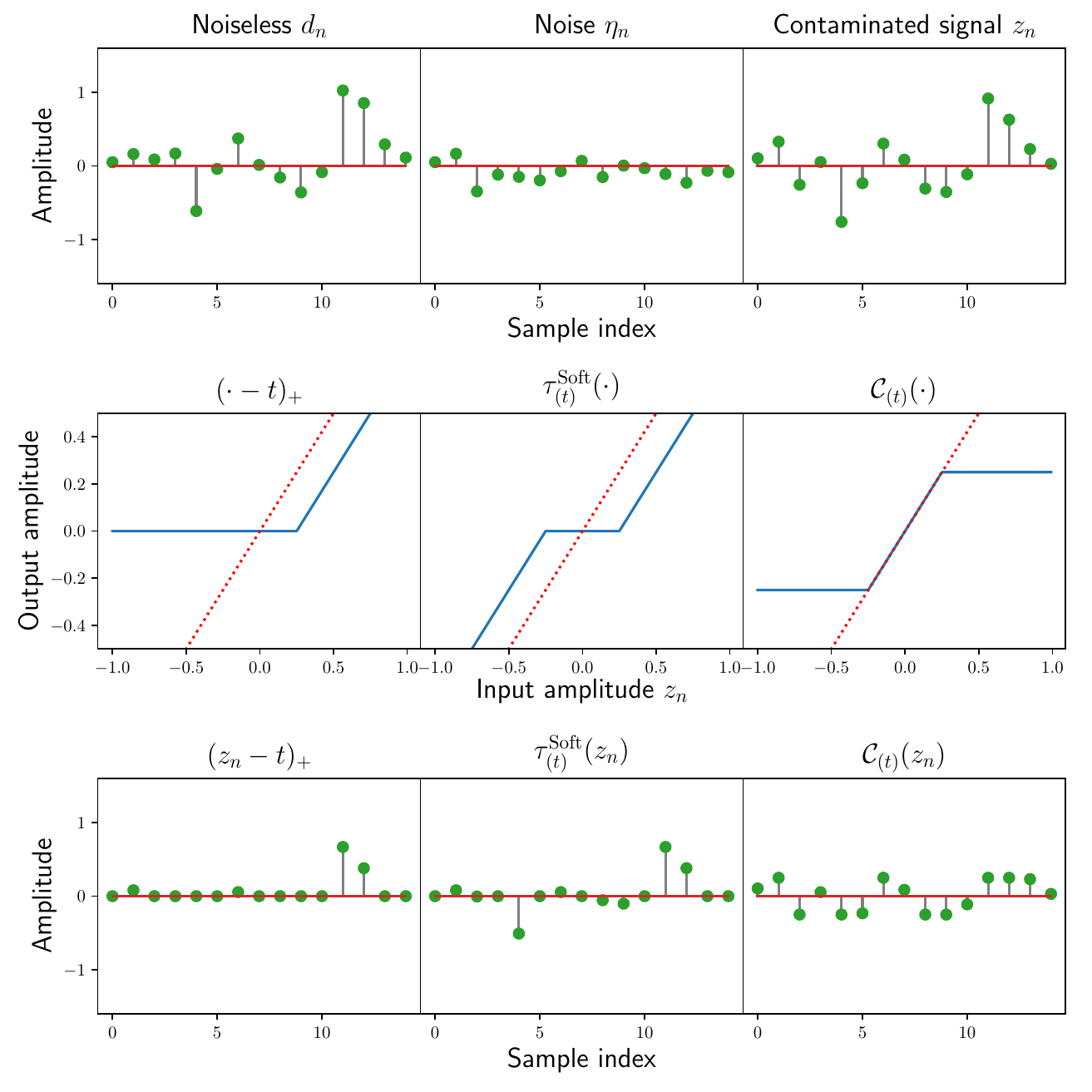}
    \caption{Signals involved in the additive noise model, input/output transfer characteristics of activation layers and estimates produced by the activation layers when applied to the noise-contaminated signal. The first row shows the signals involved in the additive noise model. The second row depicts the output amplitude of activation functions with respect to the input amplitude. Finally, the last row depicts the application of the activation functions to the noisy observation ${z}$.}
    \label{fig:fig6}
\end{figure}

For reference and further understanding, Fig.~(\ref{fig:fig6}) portrays the elements composing the noise model of Eq.~(\ref{eq:chapter4_noiseModelwaveletDomain}), the signal transfer characteristics of the ReLU, soft-shrinkage and clipping functions, and the effect that these functions have on the signal of the observed noisy detail band ${z}$.
%
%%%%%%%%%%%%%%%%%%%%%%%%%%%%%%%%
\subsubsection{\textbf{Rectified linear unit (ReLU)}}\label{sec:chapter4_ReLU}
%%%%%%%%%%%%%%%%%%%%%%%%%%%%%%%%
%
If Eq.~(\ref{eq:chapter4_logProbabSolve}) is solved for $d$ while constraining the estimator to be positive, the noiseless estimate $\hat{d}$ becomes
\begin{equation}
    \boxed{
        \hat{ {d} } =  ( {z} - t )_+
    }\ \ ,
    \label{eq:chapter4_ReLU_MAP}
\end{equation}
which is also expressed by
\begin{equation}
    ( {z} - t )_+ = 
    \begin{cases}
        z-t, &\text{if } z \geq t,\\
        \phantom{z+t}0, &\text{if } t > z .
    \end{cases}
\end{equation}
Here, the threshold level is defined by
\begin{equation}
    t = \sigma_{\eta}^2/\sigma_{d}
    .
    \label{eq:chapter4_threshold}
\end{equation}
Note that this estimator cancels the negative and low-amplitude elements of $d$ lower than the magnitude of the threshold level $t$. For example, if the signal content on the feature map is low, then $\sigma_{d} \to 0$. In such case, $t \to +\infty$ and consequently $\hat{ {d} } \to 0$. This means that the channel is suppressed. Alternatively, if the feature map has strong signal presence i.e. $\sigma_{d} \to \infty$, consequently $t \to 0$ and then $\hat{ {d} } \to ( {z} )_+ $.

A final remark is made on the modeling of functions of a CNN. It should be noted that the estimator of Eq.~(\ref{eq:chapter4_ReLU_MAP}) is analogous to the activation function of a CNN, known as \emph{rectified linear unit} (ReLU). However, in a CNN the value of $t$ would be the bias $b$ learned from the training data.
%
%%%%%%%%%%%%%%%%%%%%%%%%%%%%%%%%
\subsubsection{\textbf{Soft-shrinkage/thresholding}}\label{sec:chaper4_softShrink}
%%%%%%%%%%%%%%%%%%%%%%%%%%%%%%%%
%
If Eq.~(\ref{eq:chapter4_logProbabSolve}) is maximized in an unconstrained way, the estimate $\hat{d}$ is
\begin{equation}
    \boxed{
        \hat{ d } = \tau^\text{Soft}_{(t)}( z ) = ( z - t )_+ - (-z - t)_+
    }\ \ .
    \label{eq:chapter4softShrinkage}
\end{equation}
Here, $\tau^\text{Soft}_{(t)}( \cdot )$ denotes the soft-shrinkage/-thresholding function, which is often also written in the form
\begin{equation}
    \tau^\text{Soft}_{(t)}( z ) = 
    \begin{cases}
        z+t, &\text{if } z \geq t,\\
        \phantom{z+t}0, &\text{if } t > z  \geq -t,\\
         z-t, &\text{if } -t > z .
    \end{cases}
\end{equation}
It can be observed that the soft-threshold enforces the low-amplitude components whose magnitude is lower than the magnitude threshold level $t$ to zero. In this case, $t$ is also defined by Eq.~(\ref{eq:chapter4_threshold}). It should be noted that the soft-shrinkage estimator can also be obtained from a variational perspective~\cite{steidl2002relations}. Finally, it can be observed that the soft-shrinkage is the superposition of two ReLU functions, which has been pointed out by Fan~\emph{et~al.}~\cite{Fan2020}.
%
%%%%%%%%%%%%%%%%%%%%%%%%%%%%%%%%%%%%%%%%%%%%%%%%
\subsubsection{\textbf{Soft clipping}}\label{sec:chapter4softClip}
%%%%%%%%%%%%%%%%%%%%%%%%%%%%%%%%
%
In Section~\ref{sec:chapter4_ReLU} and Section~\ref{sec:chaper4_softShrink}, the estimate $\hat{d}$ is obtained directly from the noisy observation $z$. Alternatively, it is possible to estimate the noise $\eta$ and subtract it from $z$ akin to the residual CNNs represented by Eq.~(\ref{eq:chapter4residualLearning}). This can be achieved by solving the model
\begin{equation}
    \boxed{
        \hat{\eta} =
        z - \hat{ d} = 
        z - \tau^\text{Soft}_{( t )}( z)
    }\ \ ,
\end{equation}
which is equivalent to
\begin{equation}
    \hat{\eta} = \mathcal{C}^\text{Soft}_{( t )}(z) =  z - ( ( z - t )_+ - (-z - t)_+ ),
    \label{eq:chapter4clipping}
\end{equation}
where $\mathcal{C}_{(t)}^\text{Soft}(\cdot)$ is the \emph{soft clipping} function. Note that this function also can be expressed by
\begin{equation}
    \mathcal{C}_{(t)}^\text{Soft}(z) = 
    \begin{cases}
        \phantom{-}t, &\text{if } z \geq t,\\
        \phantom{-}z, &\text{if }  t \geq z > -t, \\
        -t, &\text{if } -t \geq z.
    \end{cases}
\end{equation}
%
%%%%%%%%%%%%%%%%%%%%%%%%%%%%%%%%
\subsubsection{\textbf{Other thresholding layers}}
%%%%%%%%%%%%%%%%%%%%%%%%%%%%%%%%
%
One of the main drawbacks of the soft-threshold activation is that it is a biased estimator. This limitation has been addressed by the hard and semi-hard thresholds, which are (asymptotically) unbiased estimators for large input values. In this section, we focus solely on the semi-hard threshold and avoid the hard variant, because is discontinuous and, therefore, not suited for models that rely on gradient-based optimization, such as CNNs.

Among the semi-hard thresholds, two notable examples are the \emph{garrote shrink} and the shrinkage functions generated by derivatives of Gaussians (DoG)~\cite{blu2007sure, zavala2022noise}. The garrote shrink function $\tau_{(\cdot)}^{\text{Gar}}(\cdot)$ is defined by
\begin{equation}
    \tau_{(t)}^{\text{Gar}} (z) = \frac{(z^2-t^2)_+}{z}
    .
\end{equation}
Furthermore, an example of a shrinkage function based on the derivative of Gaussians is given by
\begin{equation}
    \tau_{(t)}^{\text{DoG}} (z) = 
    z
    -
    \mathcal{C}_{(t)}^{\text{DoG}} (z),
\end{equation}
where the semi-hard clipping function with the derivative of Gaussians $\mathcal{C}_{(\cdot)}^{\text{DoG}} (\cdot)$ is given by 
\begin{equation}
    \mathcal{C}_{(t)}^{\text{DoG}} (z) = 
    z
    \cdot
    \exp \bigg(
        -
        \frac{z^p}{t^p}
    \bigg),
\end{equation}
in which $p$ is an even number.
\begin{figure}[h!]
    \centering
    \includegraphics[width=\columnwidth]{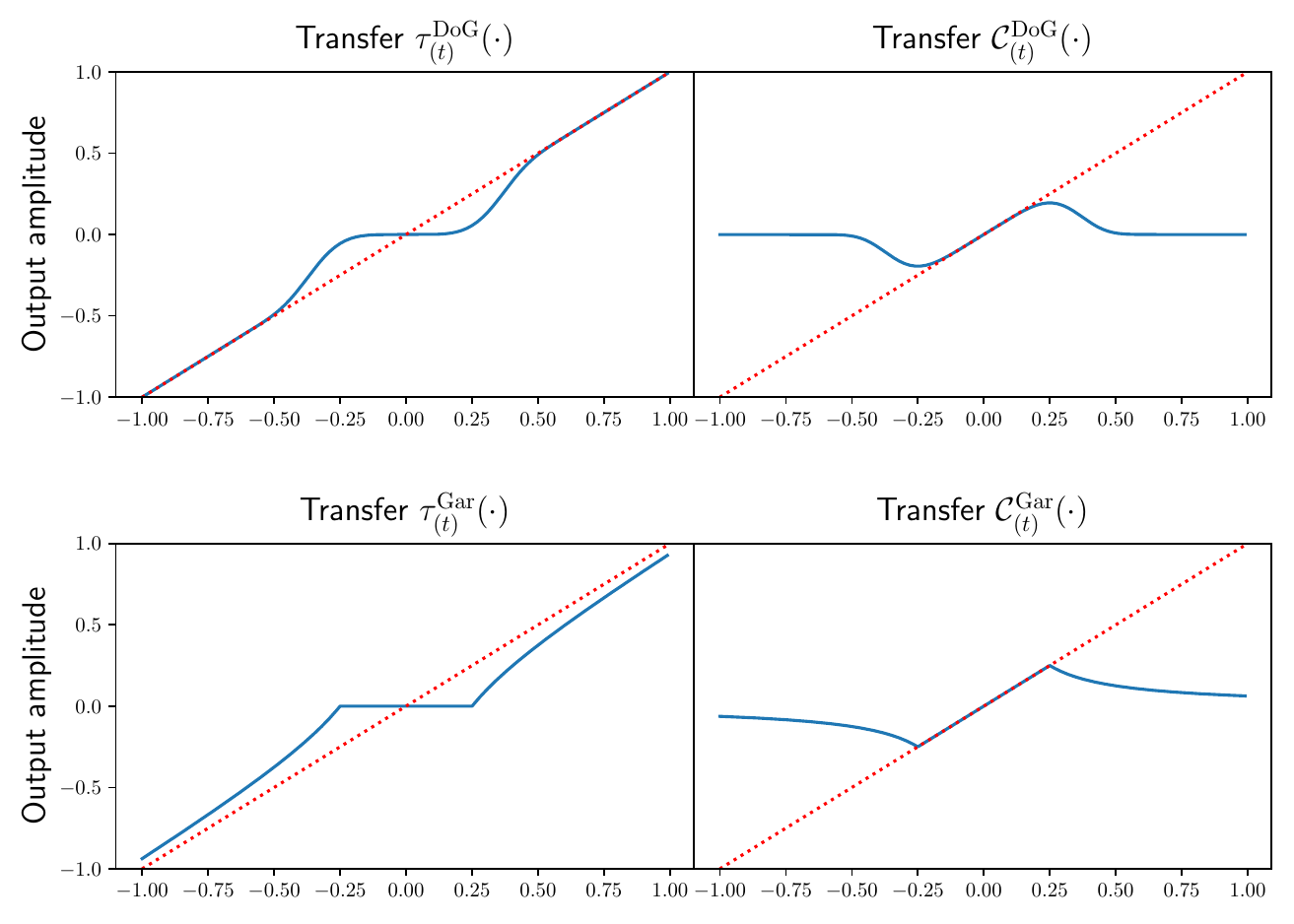}
    \caption{Transfer characteristics of the semi-hard thresholds based on the difference of Gaussians and of the garrote shrink, as well as their clipping counterparts. Note that in contrast with the soft-shrinkage and clipping functions shown in Fig.~\ref{fig:fig6}, in the semi-hard thresholds tend to unity for large values, while the semi-hard clipping functions tend to zero for large signal intensities.}
    \label{fig:fig7}
\end{figure}

The garrote and semi-hard DoG shrinkage function are shown in Fig.~\ref{fig:fig7}, as well as their clipping counterparts. Note that the shrinkage functions approximate unity for $\vert z \vert \gg t$. Therefore, they are asymptotically unbiased for large signal values.

The final thresholding function addressed in this section is the linear expansion of thresholds proposed by Blu and Luisier~\cite{blu2007sure}. This technique combines multiple thresholding functions to improve the performance. This approach is known as \emph{linear expansion of thresholds} (LET) and it is defined by
\begin{equation}
    \tau_{(\underline{t})}^\text{LET}(z) = 
    \sum_{n=0}^{N_{T}-1}
    a_n \cdot \tau_{(t_n)}(z)
    ,
\end{equation}
where $a_n$ is the weighting factor assigned to each threshold, where all weighting factor should add up to unity. 
%
%
%
%%%%%%%%%%%%%%%%%%%%%%%%%%%%%%%%
\section{\textbf{Bridging the gap between signal processing and CNNs: Deep convolutional framelets and shrinkage-based CNNs} }~\label{sec:chapter4_tdcfAndShrinkage}
%%%%%%%%%%%%%%%%%%%%%%%%%%%%%%%%
%
This section addresses the theoretical operation of noise-reduction convolutional neural networks based on ReLUs and shrinkage/thresholding functions. The first part of this section describes the theory of deep convolutional framelets~\cite{ye2018deep}, which is the most extensive study on the operation of encoding-decoding ReLU-based CNNs up to this moment. Afterwards, the section concentrates on the operation of networks which use shrinkage functions instead of ReLUs~\cite{mentl2017noise, Fan2020, zavala2022noise}, with the aim of mimicking well-established denoising algorithms~\cite{chang2000adaptive}. Finally, the last part of this section addresses the connections between both methods and additional links between convolutional neural networks and signal processing.
%
%
%%%%%%%%%%%%%%%%%%%%%%%%%%%%%%%%
\subsection{\textbf{Theory of deep convolutional framelets} }~\label{sec:chapter4_deepConvolutionalFramelets}
%%%%%%%%%%%%%%%%%%%%%%%%%%%%%%%%
%
The theory of deep convolutional framelets~\cite{ye2018deep} describes the operation of encoding-decoding ReLU-based CNNs. Its most relevant contributions are as follows. (1) To establish the equivalence of framelets and the convolutional layers of CNNs. (2) The theory of deep convolutional framelets provides the conditions to preserve the signal integrity within a ReLU CNN. (3) Explain how ReLU and convolution layers reduce noise within an encoding-decoding CNN.

The similarity between framelets and the encoding and decoding convolutional filters can be observed when comparing Eqs.~(\ref{eq:chapter4encoder}),~(\ref{eq:chaper4decoder}) with Eqs.~(\ref{eq:chapter4forFram}),~(\ref{eq:chapter4invFram}), where it becomes visible that the convolution structure of encoding-decoding CNNs is analogous to the forward and inverse framelet decomposition.

Regarding the signal reconstruction characteristics, the theory of deep convolutional framelets~\cite{ye2018deep} states the following. First, in order to be able to recover an arbitrary signal $\mathbf{y} \in \mathbb{R}^N$, the number of output channels of a convolution layer with ReLU activation should \emph{at least} duplicate the number of input channels. Second, the encoding convolution kernel $\mathbf{K}$ should be composed of pairs of filters with opposite phase. These two requirements ensure that any negative and positive values propagate through the network. Under these conditions, the encoding and decoding convolution filters $\mathbf{K}$ and $\tilde{\mathbf{K}}$ should comply with
\begin{equation}
    \boxed{
        \mathbf{y} = 
            \tilde{\mathbf{K}}^\intercal
            (
                \mathbf{K}
                \mathbf{y}
            )_+
        \cdot
        c
    }\ \ .
    \label{eq:frameletRedundancy}
\end{equation}
It can be noticed that Eq.~(\ref{eq:frameletRedundancy}) is an extension of Eq.~(\ref{eq:undecimatedFramelet}), which describes the reconstruction characteristics of tight framelets. From this point, we refer to convolutional kernels compliant with Eq~(\ref{eq:frameletRedundancy}) as \emph{phase-complementary tight framelets}. As a final remark, it should be noted that a common practice in CNN designs is also to use ReLU non-linearities in the decoder, in such case the phase-complementary tight-framelet condition can still be met as long as the pixels $y\in\mathbf{y}$ comply with $y\geq0$, which is equivalent to
\begin{equation}
    \boxed{
        \mathbf{y} = 
        (
            \mathbf{y}
        )_+
        =
        \Big(
            \tilde{\mathbf{K}}^\intercal
            (
                \mathbf{K}
                \mathbf{y}
            )_+
            \cdot
            c
        \Big)_+
    }\ \ .
    \label{eq:frameletRedundancyDec}
\end{equation}
It can be observed that the relevance of the properties defined in Eqs.~(\ref{eq:frameletRedundancy}) and (\ref{eq:frameletRedundancyDec}) is that they ensure that a CNN can propagate any arbitrary signal, which is important to avoid any distortions (such as image blur) in the processed images.

An additional element of the theory of deep convolutional framelets regarding the reconstruction of the signal, is to show that conventional pooling layers (e.g. average pooling) discard high-frequency information of the signal, which effectively blurs the processed signals. Furthermore, Ye~\emph{et al.}~\cite{ye2018deep} have demonstrated that this can be fixed by replacing the conventional up/down-sampling layers by reversible operations, such as the discrete wavelet transform. To exemplify this property, we refer to Fig.~\ref{fig:fig4}. If only an average pooling layer followed by an up-sampling stage would be applied, the treatment of the signal would be equivalent to the low-frequency branch of the DWT. Consequently, only the low-frequency spectrum of the signal would be recovered and images processed with that structure would become blurred. In contrast, if the full forward and inverse wavelet transform of Fig.~\ref{fig:fig4} is used for up- and down-sampling, it is possible to reconstruct any signal, irrespective of its frequency content.
%
%%%%%%%%%%%%%%%%%%%%%%%%%%%%%%%%
% Textbox Deep convolutional framelets
%%%%%%%%%%%%%%%%%%%%%%%%%%%%%%%%
%
\begin{figure*}
    \begin{tcolorbox}[ colback=red!10!blue!2, colframe=red!80!blue,
        title=Fitting low-rank approximation in ReLU CNNs, width=\textwidth]
        \begin{multicols}{2}
            \emph{From low-rank approximation to an encoding-decoding CNN.} In order to further understand the analogy between CNNs and low-rank approximation established by the theory of deep convolutional framelets, we can use as starting point the definition of singular value decomposition, which is expressed in Eq.~(\ref{eq:chapter4svdVector}), by
            \begin{equation*}
                \mathbf{y} = \sum_{n=0}^{N_\text{SV}-1}
                (\underline{u}_n \underline{v}_n^\intercal)
                \cdot
                \underline{\sigma}[n]
                .
            \end{equation*}
            Given that left and right singular vector pairs $\underline{u}_n\underline{v}_n^\intercal$ generate an image $\mathbf{D}[n]$, then Eq.~(\ref{eq:chapter4svdVector}) can be rewritten to
            \begin{equation}
                \mathbf{y} = \sum_{n=0}^{N_\text{SV}-1} 
                \mathbf{D}[n]
                \cdot
                \underline{\sigma}[n],  
            \end{equation}
            where tensor $\mathbf{D}=
             \begin{pmatrix}
                    (\underline{u}_0\underline{v}_0^\intercal)
                        \dots
                    (\underline{u}_{N_\text{SV}-1}\underline{v}_{N_\text{SV}-1}^\intercal)
                \end{pmatrix}^\intercal
            $ contains the products of the left and right singular vectors and has dimensions $( N_\text{SV} \times 1 \times M\times N )$. Furthermore, the equation can be further reformulated to
            \begin{equation}
                \mathbf{y} =
                \tilde{\mathbf{K}}^\intercal
                \mathbf{D}
                ,
                \label{eq:chapter4svdVector2}    
            \end{equation}
            in which
            $   \tilde{\mathbf{K}}^\intercal 
                = 
                \begin{pmatrix}
                    (\underline{\sigma}[0])
                        \dots
                    (\underline{\sigma}[N_{\text{SV}-1}])
                \end{pmatrix}
            $, where the brackets of the $(1\times 1)$ filters have been excluded for simplicity. In addition, it is now assumed that it is desirable to perform low-rank approximation of signal $ \mathbf{y} $ based on the reformulation of Eq.~(\ref{eq:chapter4svdVector2}). If we assume that $ {\mathbf{D} } \in \mathbb{R}^N_{ \geq 0 }$, then the low-rank approximation can be expressed by
            \begin{equation}
                \hat{\mathbf{y}} = 
                \tilde{\mathbf{K}}^\intercal
                (
                    \mathbf{D}
                    +
                    \underline{b}
                )_+
                ,
                \label{eq:chapter4svdVector3}
            \end{equation}
            in which the values $\underline{b}$ are set to zero for the channels of~$\mathbf{D}$ that have high contributions to the image content. Conversely, the channels of $\mathbf{D}[n]$ with less perceptual relevance are then cancelled by assigning large negative values to the corresponding entries of $\underline{b}$. As final reformulation, we can assume that the basis images $\mathbf{D}$ are the result of decomposing the input image $\mathbf{y}$ with a set of convolution filters i.e. $\mathbf{D} = \mathbf{K}\mathbf{y}$, this transforms Eq.~(\ref{eq:chapter4svdVector3}) into
            \begin{equation}
                \boxed{
                    \hat{\mathbf{x}} =
                    \tilde{\mathbf{K}}^\intercal
                    (
                        \mathbf{K}
                        \mathbf{y}
                        +
                        \underline{b}
                    )_+
                }\ \ .
                \label{eq:chapter4svdVector4}
            \end{equation}
            Here, it is visible that Eq.~(\ref{eq:chapter4svdVector4}) is analogous to the encoding-decoding architecture defined in Eqs.~(\ref{eq:chapter4encodingDecoding})~to~(\ref{eq:chaper4decoder}) and the encoder and decoder filters are akin to the framelet formulation of Section~\ref{sec:chapter4framelets}. Note that Eq.~(\ref{eq:chapter4svdVector4}) assumes that the entries $ {\mathbf{D} } = \mathbf{K}\mathbf{y}$ are positive, which may be not always true. In this situation, tensor ${\mathbf{D} }$ requires redundant channels in which their respective phases are inverted to avoid the signal loss. Furthermore, it should also be noticed that in a CNN, the bias/threshold level is not inferred from the statistics of the feature maps, but learned from the data presented to the network during training.

            \quad \emph{Multi-layer designs.} It should be noted that CNNs contain multiple layers, which recursively decompose/reconstruct the signal. This may pose an advantage with respect to conventional low-rank approximation algorithms for a few reasons. First, the data-driven nature of CNNs allows to learn the basis functions which optimally decompose and suppress noise in the signal. Second, since networks are deep, the incoming signal is recursively decomposed and sparsified. This multi-decomposition scheme is very similar to the designs used in noise-reduction algorithms based on framelets. It can be noted that the recursive sparsifying principles have been observed in the past in methods such as the (learned) iterative soft-thresholding algorithm~\cite{daubechies2004iterative, gregor2010learning} as well as convolutional sparse coding.  In fact, convolutional sparse-coding approach, which also has been used for interpreting the operation of CNNs~\cite{papyan2017convolutional}

            \quad \emph{What about practical implementations?} When training a CNN, the parameters of the model (i.e. $\mathbf{K}$, $\tilde{\mathbf{K}}^\intercal$ and $\underline{b}$) are updated to reduce the loss between the processed noisy signal and the ground truth, which does not  warranty that the numerical values of the convolution filters and biases of the trained model comply with the assumptions performed here. This is because CNNs do not have mechanisms to enforce that filters have properties such as sparsity or perfect reconstruction and negative values for the biases. Consequently, CNNs may not necessarily perform a low-rank approximation of the signal, although the mathematical formulation of the low-rank approximation and the single-layer encoding-decoding are similar. Hence, the analysis presented here should be treated as insight on the mathematical formulation and/or potential properties that can be enforced for specific applications and not as a literal description of what trained models do. 
        \end{multicols}
    \end{tcolorbox}
\end{figure*}
\begin{figure*}%[t!]
    \centering
    \includegraphics[width=\textwidth]{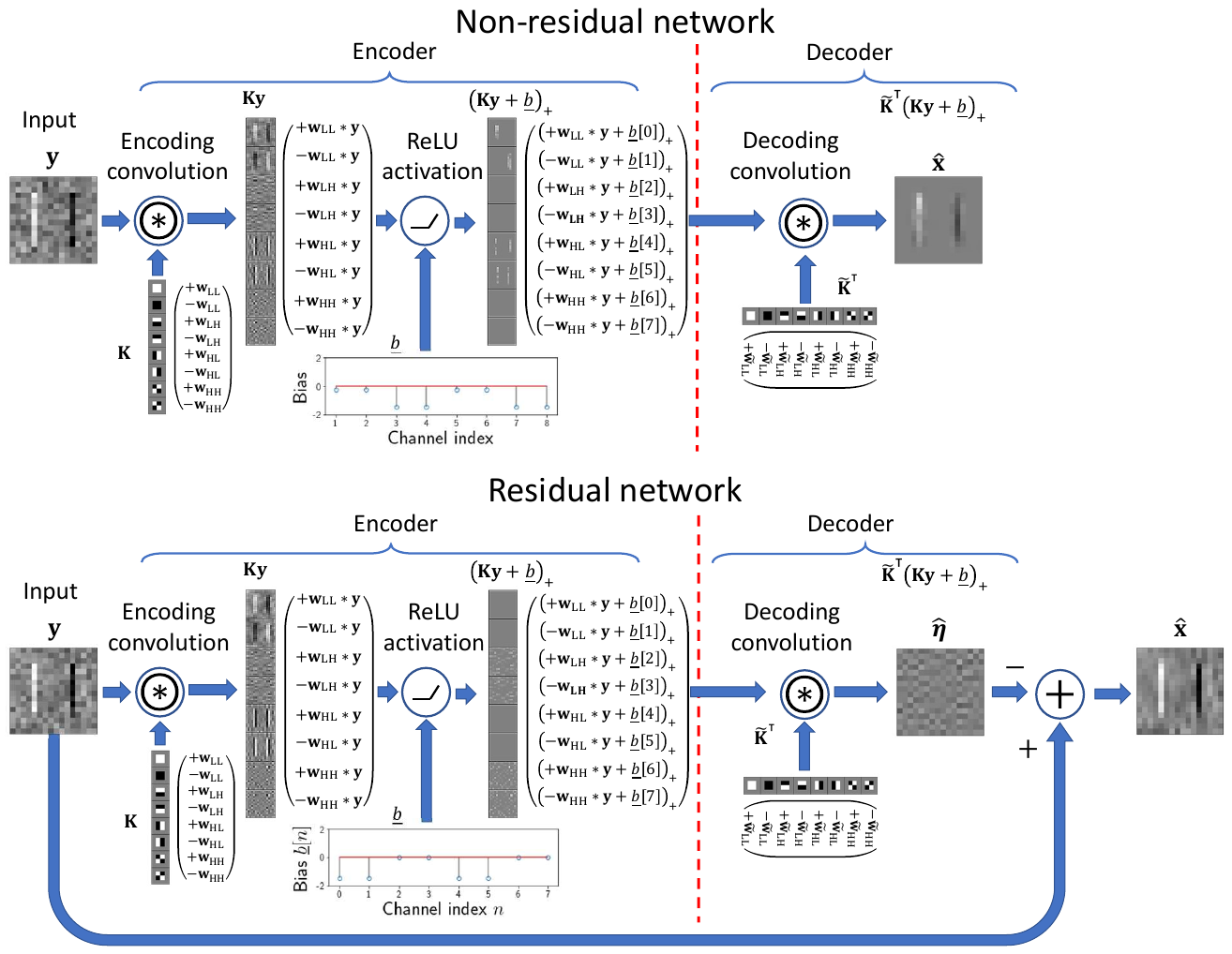}
    \caption{Operation of a simplified denoising (non-) residual ReLU CNN according to the theory of deep convolutional framelets (TDCF). In the figure, the noisy observation $\mathbf{y}$ is composed by two vertical bars plus uncorrelated Gaussian noise. Furthermore, for this example, the encoding and decoding convolution filters ($\mathbf{K}$ and $\tilde{\mathbf{K}}$, respectively) are the Haar basis of the 2D discrete wavelet transform and its phase-inverted counterparts. Given the content of the image, the image in the decomposed domain $\mathbf{K}\mathbf{y}$ produces only a weak activation for the vertical and diagonal filters ($\mathbf{w}_\textbf{LH}$ and $\mathbf{w}_\textbf{HH}$, respectively) and those feature maps contain mainly noise. In the case of the non-residual network, the ReLUs and biases suppress the channels with low activation (see column $(\mathbf{K}\mathbf{y} + \underline{b})_+$), which is akin to the low-rank approximation. In contrast, in the residual example, the channels with image content are suppressed, while preserving the uncorrelated noise. Finally, the decoding section reconstructs the noise-free estimate $\tilde{\mathbf{x}}$ for the non-residual network or the noise estimate $\hat{\boldsymbol\eta}$ for the residual example, where it is subtracted from $\mathbf{y}$ to compute the noiseless estimate $\hat{\mathbf{x}}$.}    \label{fig:fig8}
\end{figure*}

The ultimate key contribution of the theory of deep convolutional framelets is the explanation of the operation of ReLU-based noise-reduction CNNs. For the non-residual configuration, ReLU CNNs perform the following operations. (1) The convolution filters decompose the incoming signal into a sparse multi-channel representation. (2) The feature maps which are uncorrelated to the signal, contain mainly noise. In this case, the bias and the ReLU activation cancel the noisy feature maps in a process analogous to the MAP estimate shown in Section~\ref{sec:chapter4_ReLU}. (3) The decoder reconstructs the filtered image. Note that this process is analogous to the low-rank decomposition described in Section~\ref{sec:chapter4SVD}. In the case of \emph{residual} networks, the CNN learns to estimate the noise, which means that in that configuration the ReLU non-linearities suppress the channels with high activation.

A visual example of the low-rank approximation in ReLU CNNs is shown in Fig.~\ref{fig:fig8}, which illustrates the operation of an idealized single-layer encoding-decoding ReLU CNN operating both, in residual and non-residual way. It can be noted The ReLU activation suppresses specific channels in the sparse decomposition provided by the encoder, thereby preserving the low-rank structures in the non-residual network. Alternatively, in the residual example, the ReLUs eliminate the feature maps with high activation, which results in a noise estimate that is subtracted from the input to estimate the noiseless signal.
%
%
%%%%%%%%%%%%%%%%%%%%%%%%%%%%%%%%
\subsection{\textbf{Shrinkage and clipping-based CNNs} }\label{sec:chapter4shrinkageAndClipping}
%%%%%%%%%%%%%%%%%%%%%%%%%%%%%%%%
%
Just as ReLU networks, the encoder of shrinkage networks ~\cite{mentl2017noise, Fan2020, zavala2022noise} separates the input signal in a multi-channel representation. As a second processing stage, the shrinkage networks estimate the noiseless encoded signal by cancelling the low-amplitude pixels in the feature maps in a process akin to the MAP estimate of Section~\ref{sec:chaper4_softShrink}. As final step, the encoder reconstructs the estimated noiseless image. Note that the use of shrinkage functions reduces the number of channels required by ReLU counterparts to achieve perfect signal reconstruction, because the shrinkage activation preserves positive and negative values, while ReLU only preserves the positive part of the signal.

As shown in Section~\ref{sec:chapter4residualcnns}, in residual learning, a given encoding-decoding network estimates the noise signal $\boldsymbol\eta$, so that it can be subtracted from the noisy observation $\mathbf{y}$ to generate the noiseless estimate $\hat{\mathbf{x}}$. As shown in Section~\ref{sec:chapter4softClip}, in the framelet domain this is achieved by preserving the low-amplitude values of the feature maps by clipping the signal. Therefore in residual networks, the shrinkage functions can be explicitly replaced by clipping activations.

Visual examples of the operation of a single-layer shrinkage and clipping networks are presented in Fig.~\ref{fig:fig8}, where it can be noted that the operation of shrinkage and clipping networks is analogous to their ReLU counterparts, with the main difference that shrinkage and clipping networks do not require phase-complements in the encoding and decoding layers as ReLU-based CNNs do.
\begin{figure*}[t!]
    \centering
    \includegraphics[width=\textwidth]{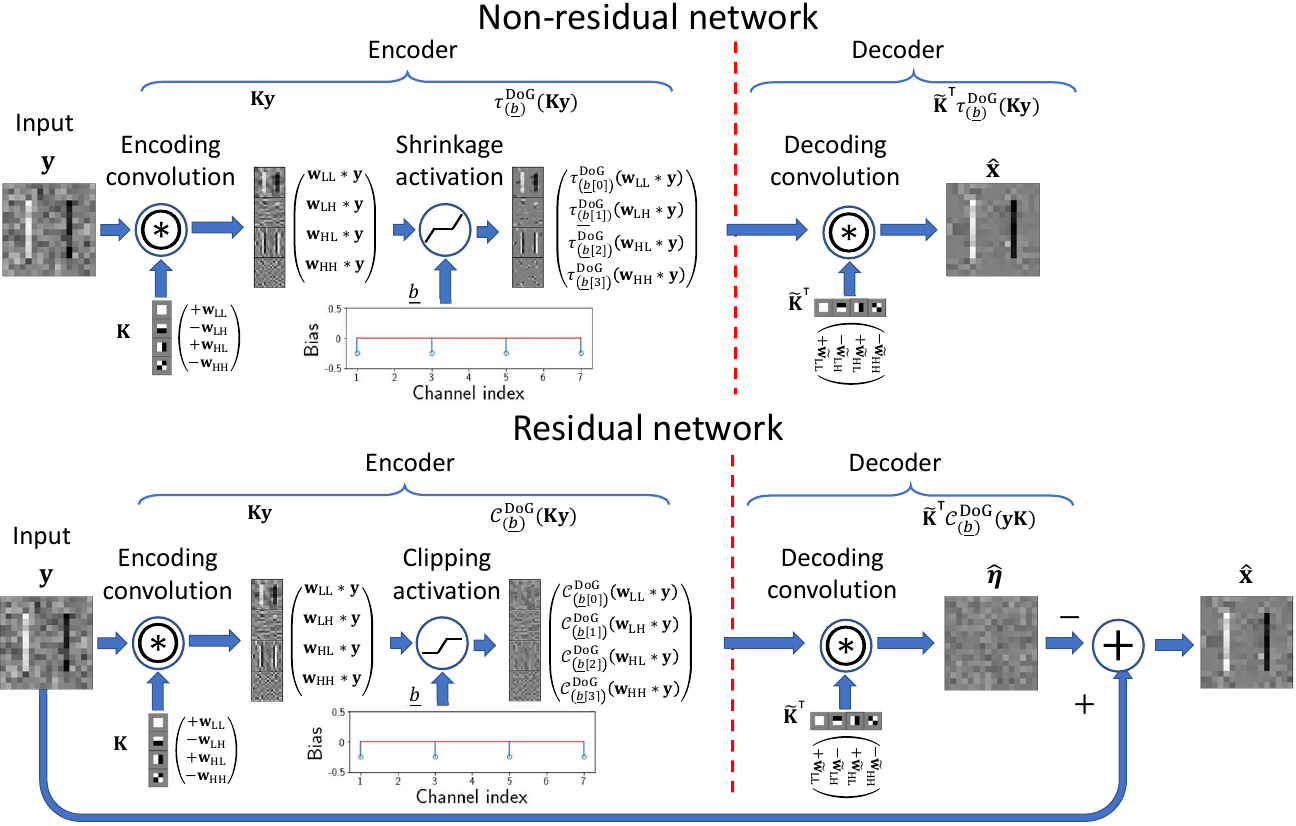}
    \caption{Operation of denoising in shrinkage and clipping networks. In the non-residual configuration, the noisy signal $\mathbf{y}$ is decomposed by a set of convolution filters, which for this example are the 2D Haar basis functions of the discrete wavelet transform ($\mathbf{K}\mathbf{y}$). As a second step, the semi-hard shrinkage produces a MAP estimate of the noiseless detail bands/feature maps ($\tau_{(\underline{b})}^\text{DoG}(\mathbf{K}\mathbf{y})$). As third and final step, the decoder maps the estimated noiseless encoded signal to the original image domain. In the residual network, the behavior is similar, but the activation layer is a clipping function which performs a MAP estimate of the noise in the feature maps, which is reconstructed by the decoder to generate the noise estimate $\hat{\boldsymbol\eta}$. After reconstruction, the noise estimate is subtracted from the noisy observation $\mathbf{y}$ to generate the noise-free estimate $\tilde{\mathbf{x}}$.}
    \label{fig:fig9}
\end{figure*}
%
%
%%%%%%%%%%%%%%%%%%%%%%%%%%%%%%%%
% Textbox Network depth
%%%%%%%%%%%%%%%%%%%%%%%%%%%%%%%%
%
\begin{figure*}
    \begin{tcolorbox}[ colback=red!10!blue!2, colframe=red!80!blue,
        title=Network depth, width=\textwidth]
        \begin{multicols}{2}
            \emph{The relationship between network depth and low-rank approximation}. It should be noted that one of the key elements of CNNs is the network depth, which we address in this section. To illustrate the effect of network depth, assume an arbitrary $N$-layer encoding-decoding CNN, in which the encoding layers are defined by
            \begin{equation}
                \begin{aligned}
                \mathbf{\mathbf{E}}_0 = &(
                    \mathbf{K}_0
                    \mathbf{y}
                    +
                    \underline{b}_0)_+, \\
                \mathbf{\mathbf{E}}_1 = &(
                    \mathbf{K}_1
                    \mathbf{\mathbf{E}}_0
                    +
                    \underline{b}_1)_+,\\
                \mathbf{\mathbf{E}}_2 = &(
                    \mathbf{K}_2
                    \mathbf{\mathbf{E}}_1
                    +
                    \underline{b}_2)_+,\\
                    \vdots \\
                \mathbf{\mathbf{E}}_{N-1} = &(
                    \mathbf{K}_{N-1}
                    \mathbf{\mathbf{E}}_{N-2}
                    +
                    \underline{b}_{N-1})_+,\\
                \end{aligned}
            \end{equation}
            \begin{equation}
                \boxed{\mathbf{\mathbf{E}}_{n} = (
                    \mathbf{K}_{n}
                    \mathbf{\mathbf{E}}_{n-1}
                    +
                    \underline{b}_{n})_+
                    }\ \ .
            \end{equation}
            Here, $\mathbf{\mathbf{E}}_n$ represents the encoded signal at the $n$-th decomposition level, while $\mathbf{K}_n$, $\underline{b}_n$ are the convolution weights and biases for the $n$-th encoding layer, respectively. As addressed in Sections~\ref{sec:chapter4_ReLU} and~\ref{sec:chapter4_deepConvolutionalFramelets}, the role of the ReLU activations is to enforce sparsity and non-negativity, which can be interpreted as the process of suppressing non-informative bases in the low-rank approximation algorithm. Consequently, every encoded signal $\mathbf{E}_n$, is an encoded \emph{sparsified} version of the signal $\mathbf{E}_{n-1}$. In order to recover the signal, we apply the decoder part of the CNN, given by 
            \begin{equation}
                \begin{aligned}
                \tilde{\mathbf{\mathbf{E}}}_{N-1} = &(
                    \tilde{\mathbf{K}}^\intercal_{N-1}
                    \mathbf{\mathbf{E}}_{N-1}
                    +
                    \tilde{\underline{b}}_{N-1})_+,\\
                \vdots \\
                \tilde{\mathbf{\mathbf{E}}}_1 =  &(
                    \tilde{\mathbf{K}}^\intercal_2
                    \tilde{\mathbf{E}}_2
                    +
                    \tilde{\underline{b}}_2)_+,\\
                \tilde{\mathbf{\mathbf{E}}}_0 =  &(
                    \tilde{\mathbf{K}}^\intercal_1
                    \tilde{\mathbf{E}}_1
                    +
                    \tilde{\underline{b}}_1)_+,\\
                \hat{\mathbf{x}} = &(
                    \tilde{\mathbf{K}}^\intercal_0
                    \tilde{\mathbf{E}}_0
                    +
                    \tilde{\underline{b}}_0)_+,
                \end{aligned}
                \label{eq:depthDecoder}
            \end{equation}
\begin{equation}
    \boxed{
    	\tilde{\mathbf{E}}_{n-1}
         =
         (
            \tilde{\mathbf{K}}^\intercal_n
            \tilde{\mathbf{E}}_{n}
             +
             \tilde{\underline{b}}_n
        )_+
    }\ \ .
    \label{eq:chaper4decoderRelu}
\end{equation}
            
            Here, $\hat{\mathbf{x}}$ is the low-rank estimate/denoised version of the input signal $\mathbf{y}$, while $\tilde{\mathbf{E}}_n$, $\tilde{\mathbf{K}}_n^\intercal$, $\tilde{\underline{b}}_n$ are the decoded signal components at the $n$-th composition level and the decoder convolution weights and biases for the $n$-th layer, respectively. In Eq.~(\ref{eq:chaper4decoderRelu}) every decoded signal $\tilde{\mathbf{E}}_n$ is the low-rank estimate of the encoded layer $\mathbf{\mathbf{E}}_{(n-1)}$. It should be noted that the activation of each of the decoder layers $(\cdot+\tilde{\underline{b}}_n)_+$ can further enforce sparsity on the low-rank estimates $\tilde{\mathbf{\mathbf{E}}}_{(n-1)}$. 
            
            \quad \emph{Summary}. In conclusion, the mathematical formulation of deep networks is analogous to a recursive data-driven low-rank approximation, where the input to the successive encoding-decoding pairs is the low-rank approximated encoded signal generated by the encoder of the previous level. Still, just as mentioned in the text box \emph{Fitting low-rank approximation in ReLU CNNs}, low-rank approximation algorithms and CNNs are similar in terms of mathematical formulation, but we cannot ensure that the values obtained during training for the encoding, decoding filters and their biases have the properties needed to ensure that a CNN is an exact recursive data-driven low-rank approximation. For example, it is possible that the filters of the encoder and decoder do not reconstruct the signal perfectly, because this may not be necessary to reduce the loss function used to optimize the network.
                        
            \quad \emph{Is it possible to impose a tighter relation between low-rank approximation and CNNs?} In specific applications where signal preservation and interpretability is required (e.g. medical imaging) it is desirable that the operation of CNNs is closer to the low-rank approximation description. In order to achieve this, the CNNs embedded in frameworks such as the convolutional analysis operator~\cite{chun2020} and FISTA-Net~\cite{Xiang21} explicitly train the filters $\mathbf{K}_n$ and $\tilde{\mathbf{K}}_n$ to have properties such as perfect signal reconstruction and sparsity. By enforcing these characteristics, the mathematical descriptions of the low-rank behavior and of CNNs are more similar and the models become inherently more interpretable and predictable on their operation.
        \end{multicols}
    \end{tcolorbox}
\end{figure*}
%
%%%%%%%%%%%%%%%%%%%%%%%%%%%%%%%%
\subsection{\textbf{Shrinkage and clipping in ReLU networks} }\label{sec:shirnkageAndClippingReLU}
%%%%%%%%%%%%%%%%%%%%%%%%%%%%%%%%
%
As addressed in Section~\ref{sec:chapter4thresh}, the soft-threshold function is the superposition of two ReLU activations. As a consequence, it is feasible that in ReLU CNNs shrinkage behavior could arise in addition to the low-rankness enforcement mentioned in Section~\ref{sec:chapter4_deepConvolutionalFramelets}. It should be noted that this only can happen if the number of channels of the encoder and decoder complies with the redundancy constraints of theory of deep convolutional framelets and if the decoder is linear. To prove this, Eq.~(\ref{eq:chapter4softShrinkage}) is reparametrized as
\begin{equation}
    \hat{\mathbf{d}} =
    \tilde{\mathbf{K}}^\intercal
    (
        \mathbf{K}
        \mathbf{y}
        +
        \underline{b}
    )_+
    ,
    \label{eq:chapter4_ReLUShrinkage}
\end{equation}
where convolution filters $\mathbf{K}$ and $\tilde{\mathbf{K}}^\intercal$ are defined by $
    \mathbf{K} =
    \begin{pmatrix}
        \begin{pmatrix}
            \mathbf{I} & -\mathbf{I}
        \end{pmatrix}
    \end{pmatrix}^\intercal
$ and $
    \tilde{\mathbf{K}}^\intercal =
    \begin{pmatrix}
        \begin{pmatrix}
            \mathbf{I} & -\mathbf{I}
        \end{pmatrix}
    \end{pmatrix}
$, respectively, and $\underline{b}=\begin{pmatrix} -t & -t \end{pmatrix}^\intercal$ represents the threshold value.

In addition to the soft-shrinkage, note that the clipping function described by Eq.~(\ref{eq:chapter4clipping}) also can be expressed by Eq.~(\ref{eq:chapter4_ReLUShrinkage}) if $
    \mathbf{K} =
    \begin{pmatrix}
        \begin{pmatrix}
            \mathbf{I} & -\mathbf{I} & \phantom{+}\mathbf{I} & -\mathbf{I}
        \end{pmatrix}
    \end{pmatrix}^\intercal
$, $
    \tilde{\mathbf{K}}^\intercal =
    \begin{pmatrix}
        \begin{pmatrix}
            \mathbf{I} & -\mathbf{I} & -\mathbf{I} & \phantom{+}\mathbf{I}
        \end{pmatrix}
    \end{pmatrix}
$ and $\underline{b}=\begin{pmatrix}0 &  \phantom{+}0 &  -t & -t \end{pmatrix}^\intercal$.
It can be noted that representing the clipping function in convolutional form requires \emph{four} times more channels than the original input signal.

It should be noted that the ability of ReLUs to approximate other signals has also been noticed Daubechies~\emph{et al.}~\cite{daubechies2022nonlinear}, who have proven that deep ReLU CNNs are universal function approximators. In addition, Ye and Sung~\cite{ye2019understanding} have demonstrated that the ReLU function is the main source of the high-approximation power of CNNs.
%
%
%%%%%%%%%%%%%%%%%%%%%%%%%%%%%%%%
\subsection{\textbf{Additional links between encoding-decoding CNNs and existing signal processing techniques} }~\label{sec:wiener}
%%%%%%%%%%%%%%%%%%%%%%%%%%%%%%%%
%
Up to this moment, it has been assumed that the operation of the encoding and decoding convolution filters is limited to map the input image to a multi-channel representation and to reconstruct it (i.e. $\mathbf{K}$ and $\tilde{ \mathbf{K} }^\intercal $ comply with $\tilde{ \mathbf{K} }^\intercal(\mathbf{K})_+ = \mathbf{I} \cdot c$). Still, it is possible that --besides decomposition and synthesis tasks-- the encoding-decoding structure also filters/colors the signal in a way that improves the image estimates. It should be noted that this implies that the perfect reconstruction encoding-decoding structure is no longer preserved. For example, considering the following linear encoding-decoding structure
\begin{equation}
    \hat{\mathbf{x}} = 
    \tilde{ \mathbf{K} }^\intercal
    (
        \mathbf{K}
        \mathbf{y}
    )
    ,
\end{equation}
which can be reduced to 
\begin{equation}
    \hat{\mathbf{x}} = \mathbf{k}*\mathbf{y}
    .
\end{equation}
Here, $\mathbf{k} = \tilde{ \mathbf{K} }^\intercal \mathbf{K} $, is optimized to reduce the distance between $ \mathbf{y}$ and the ground-truth $\mathbf{x}$. Consequently, the equivalent filter $\mathbf{k}$ can be considered a \emph{Wiener filter}. It should be noted that this text is not the first in addressing the potential Wiener-like behavior of a CNN. For example, Mohan~\emph{et al.}~\cite{Mohan2020Robust} suggested that by eliminating the bias of the convolution layers, the CNN could behave more akin to the Wiener filter and to be able to generalize better to unseen noise levels. It should be noted that by doing that the CNN can also behave akin to the switching behavior described by the theory of deep convolutional framelets, which can be described by the equation 
\begin{equation}
    (z)_+ =  
    \begin{cases}
        z, &\text{if } z \geq 0,\\
        0 &\text{if } z < 0,
    \end{cases}
    ,
\end{equation}
where $z$ is a pixel which belongs to the signal $\textbf{z} =\mathbf{k}*\mathbf{x}$. It can be observed that in contrast with the low-rank behavior described in Section~\ref{sec:chapter4_deepConvolutionalFramelets}, in this case the switching behavior is only dependent on the correlation between signal $\mathbf{x}$ and the filter $\mathbf{k}$. Consequently, if the value of $z$ is positive, its value is preserved. On the contrary, if the correlation between $\mathbf{x}$ and $\mathbf{k}$ is negative, then the value of $z$ is cancelled. Consequently, the noise reduction 
becomes independent/invariant of the noise level. It can be observed that this effect can be considered a non-linear extension of the so-called signal annihilation filters~\cite{ye2016compressive}.

It should be noticed that besides the low-rank approximation interpretation of ReLU-based CNNs, additional links to other techniques can be derived. For example, the decomposition and synthesis provided by the encoding-decoding structure is also akin to the non-negative matrix factorization (NMF)~\cite{cichocki2007nonnegative}, in which a signal is factorized as a weighted sum of positive bases. In this conception, the feature maps are the bases, which are constrained to be positive by the ReLU function. Furthermore, an additional interpretation of encoding-decoding CNNs can be obtained by analyzing them from a low-dimensional manifold representation perspective~\cite{yokota2020manifold}. Here, the convolution layers of CNNs are interpreted as two operations. On one hand, they can provide a Hankel representation. On the other hand, they provide a bottleneck which reduces the dimensionality of the manifold of image patches. It should be noticed that the Hankel-like structure that is attributed to the convolution layers of CNNs, has also been noticed by the theory of the deep convolutional framelets~\cite{ye2018deep}. Two final connection with signal-processing and CNNs is the variational formulation combined with kernel-based methods~\cite{unser2022kernel} and the convolutional sparse coding interpretation of CNNs by Papyan~\emph{et al.}~\cite{papyan2017convolutional}.
%
%
%
%
%%%%%%%%%%%%%%%%%%%%%%%%%%%%%%%%
\section{ \textbf{Mathematical analysis of relevant designs} }\label{sec:chapter4_analysis}
%%%%%%%%%%%%%%%%%%%%%%%%%%%%%%%%
%
\begin{figure}[h!]
    \centering
    \includegraphics[width=\columnwidth]{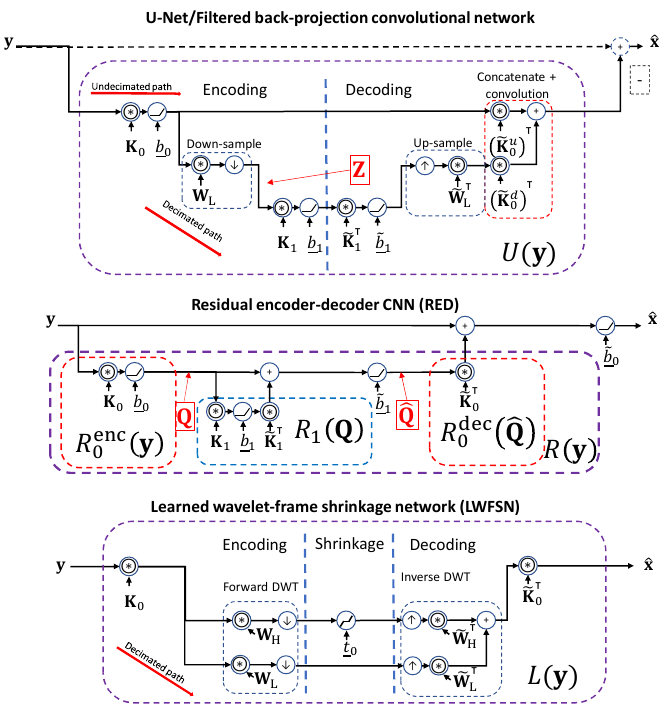}
    \caption{Simplified structure of encoding-decoding ReLU CNNs. The displayed networks are the U-Net/filtered back-projection network the encoder-decoder residual CNN (RED) and finally, the learned wavelet-frame shrinkage network. Note that for all the designs, the encoding-decoding structures are indicated by dashed blocks. It should be beared in mind that the drawings are simplified, they do not contain normalization layers, are shallow, commonly appearing dual convolutions are drawn as one layer.}
    \label{fig:fig10}
\end{figure}
In order to demonstrate the application of the principles summarized in Sections~\ref{sec:chapter4_fundamentalsOfSP} and~\ref{sec:chapter4_tdcfAndShrinkage}, this section analyzes relevant designs of ReLU and shrinkage CNNs. The analyses focus on three main aspects, which are: (1) the overall descriptions of the network architecture, (2) the signal reconstruction characteristics provided by the convolutional layers of the encoder and decoder sub-networks, and (3) the number operations $\mathcal{O}(\cdot)$ executed by the trainable parts of the network, since this will give insight on the computational requirements to execute each network and its overall complexity. 

The signal reconstruction analysis provides a theoretical indication that a given CNN design can propagate \emph{any} arbitrary signal when considering the use of ideal filters (i.e. they provide perfect reconstruction and are maximally sparse). In other words, for a fixed network architecture, there exists a selection of parameters (weights and biases) that make the neural network equal to the identity function. This result is important, because a design that cannot propagate arbitrary signals under ideal conditions, will potentially distort the signals that propagate through it \emph{by design}. Consequently, this cannot be fixed by training with large datasets and/or with the application of any special loss term. In order to understand better the signal reconstruction analysis, we provide a brief example, where it is a non-residual CNN $G(\cdot)$, where we propagate a noiseless signal $\mathbf{x}$ contaminated with noise $\boldsymbol{\eta}$, so that
\begin{equation}
    \mathbf{x} \approx G(\mathbf{x}+\boldsymbol{\eta}).
\end{equation}
Here, an ideal CNN allows to propagate \emph{any} $\mathbf{x}$, while cancelling the noise component $\boldsymbol{\eta}$, irrespective of the content of $\mathbf{x}$. If we switch our focus to an ideal \emph{residual CNN} $R(\cdot)$, it is possible to observe that
\begin{equation}
    \hat{\mathbf{x}} \approx R(\mathbf{y}) = \mathbf{y} - G(\mathbf{y}).
    \label{eq:resCNNIR}
\end{equation}
Here, $G(\cdot)$ is the encoding-decoding section of the residual network $R(\cdot)$. Consequently, it is desirable that the network $G(\cdot)$ is able to propagate the noise $\boldsymbol{\eta}$, while suppressing the noiseless signal $\mathbf{x}$, which is equivalent to
\begin{equation}
    \begin{aligned}
        \boldsymbol{\eta} \approx&  G(\mathbf{x}+\boldsymbol{\eta}).
    \end{aligned}
\end{equation}
It should be noted that in both residual and non-residual cases, there are two behaviors. On one hand, there is a signal which the network decomposes and reconstructs (almost) perfectly, and on the other hand a signal is suppressed. The signal reconstruction analysis focuses on the signals that the network can propagate or reconstruct, rather than the signal cancellation behavior. In consequence, we focus on the linear part of $G(\cdot)$ (i.e. its convolution structure), of which, according to Section~\ref{sec:chapter4_deepConvolutionalFramelets}, we assume that it handles the decomposition and reconstruction of the signal within the CNN. It should be noted that the idealized model assumed here, is only considered for analysis purposes, since practical implementations do no guarantee that this exact behavior is factually obtained. The reader is referred to Section~\ref{sec:wiener} and the text boxes \emph{Fitting low-rank approximation in ReLU CNNs} and \emph{Network depth}. 

In order to test the perfect reconstruction in non-residual CNNs, we propose the following procedure. (1) We assume an \emph{idealized} model $G(\cdot)$, where its convolution filters $\mathbf{K}_n$ and $\tilde{\mathbf{K}}_n$ comply with the phase-complementary tight framelet condition and where the biases and non-linearities suppress low-amplitude (and negative for ReLU activations) samples from the feature maps. (2) The biases/thresholds of ReLU/shrinkage CNNs are set to zero (or to infinity for clipping activations). It can be observed that this condition prevents the low-rank (or high-rank for residual models) approximation behavior of the idealized CNN. Under this circumstance, it should be possible to prove that the analyzed CNN can perfectly reconstruct \emph{any} signal. (3) The last step involves simplifying the mathematical description of the resulting model of the previous point. The mathematical simplification of the model should lead to the identity function if the model complies with the perfect reconstruction property.

To conclude the explanation on the perfect reconstruction analysis, we provide two relevant considerations. First, it can be claimed that residual networks, such as the model $R(\mathbf{y}) = \mathbf{y}-G(\mathbf{y})$ discussed in Eq.~(\ref{eq:resCNNIR}), is able to reconstruct any signal when $G(\mathbf{y})=0$ for any $\textbf{y}=\mathbf{x}+\boldsymbol{\eta}$. Still, this does not convey information about the behavior of the encoding-decoding network $G(\cdot)$, which should be able to perform perfect decomposition and reconstruction of the noise signal $\boldsymbol{\eta}$, as discussed in Eq.~(\ref{eq:resCNNIR}). To avoid this trivial solution, instead of analyzing the network $R(\cdot)$, the analysis described for non-residual models is applied to the encoding-decoding structure $G(\cdot)$, which means that the residual connection is excluded from the analysis.

The second concluding remark is that in order to distinguish the equations of the perfect signal reconstruction analysis from other models, we specify the analyzed designs of the perfect reconstruction models, in which the low-rank approximation behavior is avoided by setting the bias values to zero, with a special operator $\mathcalboondox{P}\{\cdot\}$.

For the analyses regarding the total number of operations of the trainable parameters, it is assumed that the tensors $\mathbf{K}_0$, $\tilde{\mathbf{K}}_0^\intercal$, $(\tilde{\mathbf{K}}_0^\text{u})^\intercal$, $(\tilde{\mathbf{K}}_0^\text{d})^\intercal$, $\mathbf{K}_1$ and $\tilde{\mathbf{K}}_1^\intercal$ shown in Fig.~\ref{fig:fig10} have dimensions $(C_{0} \times 1 \times N_\text{f} \times N_\text{f})$, $(1 \times C_{0} \times N_\text{f} \times N_\text{f})$, $(1 \times C_{0} \times N_\text{f} \times N_\text{f})$, $(1 \times C_{0} \times N_\text{f} \times N_\text{f})$, $(C_{1} \times C_{0} \times N_\text{f} \times N_\text{f})$, $(C_{0} \times C_{1} \times N_\text{f} \times N_\text{f})$, respectively. Here, $C_0$ and $C_1$ represent the number of channels after the first and second convolution layer, all the convolution filters are assumed to be square with $N_\text{f} \times N_\text{f}$ pixels. Furthermore, the input signal $\mathbf{x}$ has dimensions $(1 \times 1 \times N_\text{r} \times N_\text{c})$, where $N_\text{r}$ and $N_\text{c}$ denote the numbers of rows and columns, respectively.

The analyses shown for the different networks in this article have the following limitations. (1) The analyzed networks have only enough decomposition levels and convolution layers to understand their basic operation. The motivation for this simplification is to keep the analyses short and clear. Moreover, the same principles can be extended to deeper networks. Since the same single-decomposition CNNs would be recursively embedded within the given architectures. (2) The normalization layers are not considered, because they are linear operators which provide mean shifts and amplitude scaling. Consequently, for analysis purposes  it can be assumed that they are embedded in the convolution weights. (3) For every encoder convolution kernel it is assumed that there is at least one decoder filter. (4) No co-adaptation between the filters of the encoder and decoder layers is considered.

The remainder of this section shows analyses of a selection of a few representative designs. Specifically, the chosen designs are the \emph{U-Net}~\cite{ronnebergerUnet} and its residual counterpart, the \emph{filtered back-projection network}~\cite{jin2017deep}~\footnote{Matlab implementation by their authors available at \url{https://github.com/panakino/FBPConvNet}}. Additional designs analyzed here are the \emph{the residual encoder-decoder CNN}~\cite{chen2017low}~\footnote{Pytorch implementation by their authors available at \url{https://github.com/SSinyu/RED-CNN} }, as well as the \emph{learned wavelet-shrinkage network} (LWFSN)~\footnote{Pytorch implementation available at \url{https://github.com/LuisAlbertZM/demo_LWFSN_TMI} and interactive demo available at IEEE's code ocean \url{https://codeocean.com/capsule/9027829/tree/v1}. The demo also includes as reference pytorch implementations of FBPConvNet and the tight-frame U-Net.}. For reference, all the designs are portrayed in Fig.~\ref{fig:fig10}.
%
%%%%%%%%%%%%%%%%%%%%%%%%%%%%%%%%
\subsection{\textbf{U-Net/filtered back-projection network}}\label{sec:unet}
%%%%%%%%%%%%%%%%%%%%%%%%%%%%%%%%
% 
%%%%%%%%
\subsubsection{\textbf{U-Net -- overview of the design}}
%%%%%%%%
%
The first networks analyzed are the U-Net and filtered back-projection networks both of which share the encoding-decoding structure $U(\cdot)$. However, they differ in the fact that the U-Net is non-residual, while the filtered back-projection network operates in residual configuration. Therefore, the estimate of the noiseless signal $\hat{\mathbf{x}}$ from the noisy observation $\mathbf{y}$ in the conventional U-Net is achieved by
\begin{equation}
    \hat{\mathbf{x}} = U(\mathbf{y})
    ,
\end{equation}
whereas in the filtered back-projection network, $U(\cdot)$ is used in residual configuration, which is equivalent to
\begin{equation}
    \hat{\mathbf{x}} = \mathbf{y} - U(\mathbf{y}).
\end{equation}
If we now switch focus to the encoding-decoding structure of the U-Net $U(\mathbf{y})$, it can be shown that it is described by
\begin{equation}
    U( \mathbf{y} ) = U^\text{u}( \mathbf{y} )
    + 
    U^\text{d}( \mathbf{y} )
    ,
    \label{eq:chapter4unet}
\end{equation}
where $U^\text{u}( \mathbf{y} ) $ corresponds to the \emph{undecimated path} and is defined by
\begin{equation}
    U^\text{u}( \mathbf{y} ) =
    \big(\tilde{\mathbf{K}}_0^\text{u}\big)^\intercal
    \big(
        \mathbf{K}_0
        \mathbf{y} 
        +\underline{b}_0
    \big)_+
    ,
\end{equation}
while the decimated path is
\begin{equation}
    U^\text{d}( \mathbf{y} ) =
    \big(\tilde{\mathbf{K}}_0^d\big )^\intercal
    \tilde{\mathbf{W}}_\text{L}^\intercal
    f_{(2\uparrow)}
    \Big(
            \big(
                \tilde{\mathbf{K}}_1^\intercal
                (
                    \mathbf{K}_1
                    \mathbf{Z}
                    +
                    \underline{b}_1
                )_+
                +
                \tilde{\underline{b}}_1
            \big)_+
    \Big)
    .
\end{equation}
Here, signal $\mathbf{Z}$ is defined by
\begin{equation}
     \mathbf{Z} = f_{(2\downarrow)} \big(
        \mathbf{W}_\text{L}
        (
            \mathbf{K}_0
            \mathbf{y} 
            +
            \underline{b}_0
        )_+
    \big).
\end{equation}
Note the decimated path contains two nested encoding-decoding architectures, as observed by Jin~\emph{et al.}~\cite{jin2017deep}, who has acknowledged that the nested filtering structure is akin to the (learned) iterative soft thresholding algorithm~\cite{daubechies2004iterative, gregor2010learning}.
% 
%%%%%%%%
\subsubsection{\textbf{U-Net -- signal reconstruction analysis}}
%%%%%%%%
%
To prove if the U-Net can perfectly reconstruct \emph{any} signal, we assume that the biases are equal to zero, on this condition the network $\mathcalboondox{P} \{ U \}( \mathbf{y} )$ is defined by
\begin{equation}
    \mathcalboondox{P} \{ U \}( \mathbf{y} ) = 
    \mathcalboondox{P} \{ U^\text{u}  \}( \mathbf{y} ) +
    \mathcalboondox{P} \{ U^\text{d} \}( \mathbf{y} )
    ,
    \label{eq:chapter4_unetLin}
\end{equation}
where sub-network $\mathcalboondox{P} \{ U^\text{u} \}( \cdot )$ is defined by
\begin{equation}
    \mathcalboondox{P} \{ U^\text{u} \}( \mathbf{y} ) =
    \big(\tilde{\mathbf{K}}_0^\text{u}\big)^\intercal
    (
        \mathbf{K}_0
        \mathbf{y}
    )_+
    .
\end{equation}
Assuming that $\big(\mathbf{K}_0, \tilde{\mathbf{K}}_0^\text{u}\big)$ is a complementary-phase tight-framelet pair then
$
    \mathcalboondox{P}
    \{ U^\text{u} \}( \mathbf{y} )
$ is simplified to
\begin{equation}
    \mathcalboondox{P} \{ U^\text{u} \}( \mathbf{y} ) =
    \mathbf{y}
    \cdot
    c_0.
    \label{eq:chapter4_unetUndLin}
\end{equation}
Furthermore, the low-frequency path is 
\begin{equation}
    U^\text{d}( \mathbf{y} ) =
    \big(\tilde{\mathbf{K}}_0^d\big )^\intercal
    \tilde{\mathbf{W}}_\text{L}^\intercal
    f_{(2\uparrow)}\Big(
            \big(
                \tilde{\mathbf{K}}_1^\intercal
                (
                    \mathbf{K}_1
                    \mathcalboondox{P}\{\mathbf{Z}\}
                )_+
            \big)_+
        \Big)
        ,
    \label{eq:unetRecDec}
\end{equation}
Where $\mathcalboondox{P}\{\mathbf{Z}\}$ is defined by
\begin{equation}
     \mathcalboondox{P}\{\mathbf{Z}\} = f_{(2\downarrow)}
     \big(
        \mathbf{W}_\text{L}
        (
            \mathbf{K}_0
            \mathbf{y} 
        )_+
    \big).
\end{equation}
If $\mathbf{K}_1$ is a phase-complementary tight frame, we know that $
    \tilde{\mathbf{K}}_1^\intercal
    (
        \mathbf{K}_1
        \mathbf{Z}
    )_+
    =
    \mathbf{Z} \cdot c_1
$. Consequently, Eq.~(\ref{eq:unetRecDec}) becomes
\begin{equation}
    \mathcalboondox{P} \{ U^\text{d} \}( \mathbf{y} ) = 
    \big( \tilde{\mathbf{K}}_0^d \big)^\intercal
    \tilde{\mathbf{W}}_\text{L}^\intercal
    f_{(2\uparrow)}(
        f_{(2\downarrow)} \big(
            (
                \mathbf{W}_\text{L}
                \mathbf{K}_0
                \mathbf{y}
            )_+
        \big)
    )
    \cdot
    c_1
    .
\end{equation}
Here, it can be noticed that if $\mathbf{K}_0$ is a phase-complementary tight framelet, then $\mathcalboondox{P} \{ U^\text{d} \}( \mathbf{y} ) $ approximates a low-pass version of $\mathbf{y}$, or equivalently
\begin{equation}
    \mathcalboondox{P} \{ U^\text{d} \}( \mathbf{y} ) 
    \approx
    \boldsymbol{\mathcal{W}}_\text{L}
    \mathbf{y}
    \cdot
    c_1
    ,
    \label{eq:chapter4_unetDecLin}
\end{equation}
where $\boldsymbol{\mathcal{W}}_\text{L}$ is a low-pass filter. Finally, substituting Eq.~(\ref{eq:chapter4_unetUndLin}) and  Eq.~(\ref{eq:chapter4_unetDecLin}) in Eq.~(\ref{eq:chapter4_unetLin}) results in
\begin{equation}
    \mathcalboondox{P} \{ U \}( \mathbf{y} )
    \approx
    (
        \mathbf{I}
        \cdot
        c_0 
        +
        \boldsymbol{\mathcal{W}}_\text{L}
        \cdot
        c_1
    )
    \mathbf{y}
    .
    \label{eq:responseU}
\end{equation}
This result proves that the design of the U-Net cannot evenly reconstruct all the frequency of $\mathbf{y}$ unless $c_1=0$, in which case, the whole low-frequency branch of the network is ignored. Note that this limitation is inherent to its design and cannot be circumvented by training with large datasets and/or with any loss function.
%
%%%%%%%%%%%%%%%%%%%%%%%%%%%%%%%%
\subsubsection{\textbf{U-Net -- number of operations}}
%%%%%%%%%%%%%%%%%%%%%%%%%%%%%%%%
%
It can be noted that encoding filter $\mathbf{K}_0$, convolves $\mathbf{x}$ at its original resolution and maps it to a tensor with $C_0$ channels. Therefore, the number of operations $\mathcal{O}(\cdot)$ for kernel $\mathbf{K}_0$ is $\mathcal{O}( {\mathbf{K}_0} ) = C_0 \cdot N_\text{r} \cdot N_\text{c} \cdot N_\text{f}^2~[\text{FLOPS}]$ (floating-point operations). Conversely, due to the symmetry between encoder and decoder filters, $\mathcal{O}( \tilde{{\mathbf{K}}_0^\text{u}} ) = \mathcal{O}( \tilde{{\mathbf{K}}_0^\text{d}} ) = \mathcal{O}( {\mathbf{K}_0} )$. Furthermore, for this design, filter $\mathbf{K}_1$ processes the signal encoded by $\mathbf{K}_0$, which is down-sampled by a factor of one half and maps it from $C_0$ to $C_1$ channels, this results in the estimated operation cost $\mathcal{O}(\mathbf{K}_1) = \mathcal{O}(\tilde{\mathbf{K}}_1) = C_0 \cdot C_1 \cdot N_\text{r} \cdot N_\text{c} \cdot N_\text{f}^2 \cdot (2)^{-2}~[\text{FLOPs}]$. Finally, adding the contributions of filters $\mathbf{K}_0$, $\tilde{\mathbf{K}}_0^\text{u}$, $\tilde{\mathbf{K}}_0^\text{d}$, $\mathbf{K}_1$ and $\tilde{\mathbf{K}}_1$ results in 
\begin{equation}
    \mathcal{O}(U) = (3 + 2^{-1} \cdot C_1)\cdot C_0 \cdot N_\text{r} \cdot N_\text{c} \cdot N_\text{f}^2~[\text{FLOPS}]
\end{equation}
%
%%%%%%%%%%%%%%%%%%%%%%%%%%%%%%%%
\subsubsection{\textbf{U-Net -- Concluding remarks}}
%%%%%%%%%%%%%%%%%%%%%%%%%%%%%%%%
%
The U-Net/FBPConvNet is a flexible multi-resolution architecture. Still, as it has been shown, the pooling structure of this CNN may be sub-optimal for noise reduction applications because this configuration does not allow to recover the frequency information of the signal evenly. This has been noted and fixed by Han and Ye~\cite{han2018framing}, who introduced the so-called tight-frame U-Net. In which the down/up-sampling structure is replaced by the discrete wavelet transform and its inverse. This simple modification overcomes the limitations of the U-Net and improved its performance for artifact removal in compressed sensing imaging.
%
%
%%%%%%%%%%%%%%%%%%%%%%%%%%%%%%%%
\subsection{ \textbf{Residual encoder-decoder CNN} }
%%%%%%%%%%%%%%%%%%%%%%%%%%%%%%%%
%
% 
%%%%%%%%
\subsubsection{\textbf{Residual encoder-decoder CNN -- overview of the design}}
%%%%%%%%
%
The residual encoder-decoder CNN shown in Fig.~\ref{fig:fig10} consists of nested single-layer residual encoding-decoding networks. For example, in the network showcased in Fig.~\ref{fig:fig10} it is visible that network $R_1(\cdot)$ is nested into $R_0(\cdot)$. Furthermore, for this case the image estimate is given by
\begin{equation}
    \hat{\mathbf{x}} = (
        \mathbf{y}
        +
            R_0( \mathbf{y} )
        +   
        \tilde{\underline{b}}_0
    \big)_+
    ,
    \label{eq:chapter4_RED}
\end{equation}
in which $R_0(\cdot)$ is the outer residual network and $\tilde{\underline{b}}_0$ is the bias for the output layer. Note that the ReLU placed at the output layer intrinsically assumes that the estimated signal $\hat{\mathbf{x}}$ is positive.

From Eq.~(\ref{eq:chapter4_RED}), the output of the sub-network $R_0(\cdot)$ is defined by
\begin{equation}
    \mathbf{Z} = 
    R^\text{dec}_0(
        \hat{\mathbf{Q}}
    ).
\end{equation}
Here, the decoder $R^\text{dec}_0(\cdot)$ is defined by
\begin{equation}
    R^\text{dec}_0(\hat{\mathbf{Q}})
    =
    \tilde{{\textbf{K}}}_0^\intercal
    \hat{\mathbf{Q}}
    .
\end{equation}
In the above, $\hat{\mathbf{Q}}$ is the noiseless estimate of the intermediate signal $\mathbf{Q}$ and it is defined by
\begin{equation}
    \hat{\mathbf{Q}} = (
        \mathbf{Q}
        +
        R_1(
            \mathbf{Q}
        )
        +
        \tilde{\underline{b}}_1
    )_+
    ,
\end{equation}
where the network $R_1(\cdot)$ is
\begin{equation}
    R_1(\mathbf{Q}) =
    \tilde{\mathbf{K}}_1^\intercal
    \big(
        \mathbf{K}_1
        \mathbf{Q}
        +
        \underline{b}_1
    )_+
    .
\end{equation}
Furthermore, $\mathbf{Q}$ represents the signal encoded by $R_0(\cdot)$, or equivalently 
\begin{equation}
    \mathbf{Q}
    =
    R_0^\text{enc}(
        \mathbf{y}
    ),
\end{equation}
where $R_0^\text{enc}(\cdot)$ is defined by
\begin{equation}
    R^\text{enc}_0(\mathbf{y})
    =
    \textbf{K}_0
    \mathbf{y}
    .
\end{equation}
%
%%%%%%%%
\subsubsection{\textbf{Residual encoder-decoder CNN -- signal reconstruction analysis}}
%%%%%%%%
%
As mentioned earlier, the residual encoder-decoder CNN is composed by nested residual blocks, which are independently analyzed to study the reconstruction characteristics of this network. First, block $R_1(\cdot)$, is given by
\begin{equation}
    \mathcalboondox{P} \{ R_1 \}(\mathbf{Q}) = 
    \tilde{\mathbf{K}}_1^\intercal
    (
        \mathbf{K}_1
        \mathbf{Q}
    )_+
    .
    \label{eq:chapter4_RED_lin_R1}
\end{equation}
Under complementary-phase tight-frame assumptions for the pair $(\mathbf{K}_1, \tilde{\mathbf{K}}_1)$, Eq.~(\ref{eq:chapter4_RED_lin_R1}) reduces to
\begin{equation}
    \mathcalboondox{P} \{ R_1 \}(\mathbf{Q}) = 
    \mathbf{Q}
    ,
    \label{eq:chapter4_RED_lin_R1red}
\end{equation}
which shows that the encoder and decoder $R_1(\cdot)$ can approximately reconstruct any signal. Now switching to $R_0$ it can be observed that the linear part is
\begin{equation}
    \mathcalboondox{P} \{ R_0 \}( \mathbf{y} ) = 
    \tilde{\mathbf{K}}_0^\intercal
    (
        \mathbf{K}_0
        \mathbf{y}
    )_+
    .
    \label{eq:chapter4_RED_lin_R0}
\end{equation}
Just as with $R_1(\cdot)$, it is assumed that the convolution kernels are tight-framelets. Therefore Eq.~(\ref{eq:chapter4_RED_lin_R0}) becomes
\begin{equation}
    \mathcalboondox{P} \{ R_0 \}(\mathbf{y}) = 
    \mathbf{y}
    .
    \label{eq:chapter4_RED_lin_R0red}
\end{equation}
Consequently, $R_0(\cdot)$ and $R_1(\cdot)$ can reconstruct any arbitrary signal under complementary-phase tight-frame assumptions.
%
%%%%%%%%%%%%%%%%%%%%%%%%%%%%%%%%
\subsubsection{\textbf{Residual encoder-decoder CNN -- number of operations}}
%%%%%%%%%%%%%%%%%%%%%%%%%%%%%%%%
%
In this case,  all the convolution layers operate at the original resolution of image~$\mathbf{x}$. Therefore, the number of operations $O(\cdot)$ for kernel~${\mathbf{K}}_0$ and $\tilde{{\mathbf{K}}}_0$ is~$O( {{\mathbf{K}}_0} ) = O( \tilde{{\mathbf{K}}_0} ) = C_0 \cdot N_\text{r} \cdot N_\text{c} \cdot N_\text{f}^2$~[FLOPs], while ${\mathbf{K}}_1$ and $\tilde{{\mathbf{K}}}_1$ are requiring $O({\mathbf{K}}_1) = O(\tilde{{\mathbf{K}}}_1) = C_0 \cdot C_1 \cdot N_\text{r} \cdot N_\text{c} \cdot N_\text{f}^2$~[FLOPs]. By adding the contributions of both encoding-decoding pairs, the total operations for the residual encoder-decoder becomes
\begin{equation}
    \mathcal{O}(R) = 2\cdot(1 + C_1) \cdot C_0 \cdot N_\text{r} \cdot N_\text{c} \cdot N_\text{f}^2~[\text{FLOPS}]
    .
\end{equation}
%
%%%%%%%%%%%%%%%%%%%%%%%%%%%%%%%%
\subsubsection{\textbf{Residual encoder-decoder CNN -- concluding remarks}}
%%%%%%%%%%%%%%%%%%%%%%%%%%%%%%%%
%
The residual encoder-decoder network consists on a set of nested single-resolution residual encoding-decoding CNNs. The single-resolution design increases its computation cost with respect to multi-resolution designs such as the U-Net. In addition, it should be noted that the use of a ReLU as output layer of the encoder-decoder residual network forces the signal estimates to be positive, but this is not always convenient. For example in computed tomography imaging, it is common that images contain positive and negative values.
%
%%%%%%%%%%%%%%%%%%%%%%%%%%%%%%%%
\subsection{ \textbf{Learned wavelet-frame shrinkage network} }\label{sec:LWFSN}
%%%%%%%%%%%%%%%%%%%%%%%%%%%%%%%%
%
%%%%%%%%%%%%%%%%
\subsubsection{\textbf{Learned wavelet-frame shrinkage network -- description of the architecture}}
%%%%%%%%%%%%%%%%
%
The learned wavelet-frame shrinkage network is a multi-resolution architecture in which the discrete wavelet transform is used for down/up-sampling and also as part of the decomposition where shrinkage is applied. In this CNN, the noiseless estimates are produced by
\begin{equation}
    \hat{\mathbf{x}} = L(\mathbf{y}),
    \label{eq:chapter4_lwfsn0}
\end{equation}
where $L(\cdot)$ represents the encoding-decoding structure of the learned wavelet-frame shrinkage network and the encoding-decoding network $L(\cdot)$ is 
\begin{equation}
    L(\mathbf{y}) =
    L_\text{L}(\mathbf{y})
    +
    L_\text{H}(\mathbf{y})
    .
    \label{eq:chapter4_lwfsn}
\end{equation}
Here, the high-frequency path is given by
\begin{equation}
    L_\text{H}(\mathbf{y}) =
    \tilde{\mathbf{K}}_0^\intercal
    \tilde{\mathbf{W}}_\text{H}^\intercal
    f_{(2\uparrow)} \big(
        \tau_{(\underline{t}_0)}^\text{LET}\big(
            f_{(2\downarrow)} \big(
                \mathbf{W}_\text{H}
                \mathbf{K}_0
                \mathbf{y}
            \big)
        \big)
    \big)
    .
    \label{eq:chapter4_lwfsn_high}
\end{equation}
Note that in this design the encoder leverages the filter $\mathbf{W}_\text{H}$ for generating a sparse signal prior to the shrinkage stage, i.e. $\tau_{(\underline{t}_0)}^\text{LET}\big(
    f_{(2\downarrow)} \big(
        \mathbf{W}_\text{H}
        \mathbf{K}_0
        \mathbf{y}
    \big)
\big)$. Meanwhile, the low-frequency path $L_\text{L}(\cdot)$ is
\begin{equation}
    L_\text{L}(\mathbf{y}) =
    \tilde{\mathbf{K}}_0^\intercal
    \tilde{\mathbf{W}}_\text{L}^\intercal
    f_{(2\uparrow)} \big(
        f_{(2\downarrow)} \big(
            \mathbf{W}_\text{L}
            \mathbf{K}_0
            \mathbf{y}
        \big)
    \big)
    .
    \label{eq:chapter4_lwfsn_low}
\end{equation}
%
%%%%%%%%
\subsubsection{\textbf{Learned wavelet-frame shrinkage network -- signal reconstruction analysis}}
%%%%%%%%
%
When analyzing the signal propagation of the learned wavelet-frame shrinkage network, we set the threshold level $\underline{t_0}=0$. This turns Eq.~(\ref{eq:chapter4_lwfsn}) turns into
\begin{equation}
    \mathcalboondox{P} \{ L \}(\mathbf{y}) =
    \mathcalboondox{P} \{ L_\text{L} \}(\mathbf{y})
    +
    \mathcalboondox{P} \{ L_\text{H} \}(\mathbf{y}).
    \label{eq:chapter4_lwfsn_lin}
\end{equation}
Here, $\mathcalboondox{P} \{ L_\text{H} \}( \cdot )$ is defined by 
\begin{equation}
    \mathcalboondox{P} \{ L_\text{H} \}(\mathbf{y}) =
    \tilde{\mathbf{K}}_0^\intercal
    \tilde{\mathbf{W}}_\text{H}^\intercal
    f_{(2\uparrow)} \big(
        f_{(2\downarrow)} \big(
            \mathbf{W}_\text{H}
            \mathbf{K}_0
            \mathbf{y}
        \big)
    \big)
    ,
    \label{eq:chapter4_lwfsn_high_lin}
\end{equation}
while the low-frequency path $\mathcalboondox{P} \{ L_\text{L} \}( \cdot )$ is mathematically described by
\begin{equation}
    \mathcalboondox{P} \{ L_\text{L} \}(\mathbf{y}) =
    \tilde{\mathbf{K}}_0^\intercal
    \tilde{\mathbf{W}}_\text{L}^\intercal
    f_{(2\uparrow)} \big(
        f_{(2\downarrow)} \big(
            \mathbf{W}_\text{L}
            \mathbf{K}_0
            \mathbf{y}
        \big)
    \big)
    .
    \label{eq:chapter4_lwfsn_low_lin}
\end{equation}
Substituting Eq.~(\ref{eq:chapter4_lwfsn_high_lin}) and Eq.~(\ref{eq:chapter4_lwfsn_low_lin}) in Eq.~(\ref{eq:chapter4_lwfsn_lin}) results in the equation
\begin{equation}
    \mathcalboondox{P} \{ L\}(\mathbf{y}) =
    \tilde{\mathbf{K}}_0^\intercal
    \tilde{\mathbf{W}}^\intercal
    f_{(2\uparrow)} \big(
        f_{(2\downarrow)} \big(
            \mathbf{W}
            \mathbf{K}_0
            \mathbf{y}
        \big)
    \big)
    .
    \label{eq:chapter4_lwfsn_lin2}
\end{equation}
For the discrete wavelet transform, it holds that $\mathbf{Q} =
\tilde{\mathbf{W}}^\intercal
f_{(2\uparrow)}
\big(
    f_{(2\downarrow)}
    \big(
        \mathbf{{W}}
        \mathbf{Q}
    \big)
\big)
$. Consequently, Eq.~(\ref{eq:chapter4_lwfsn_lin2}) is simplified to
\begin{equation}
    \mathcalboondox{P} \{ L\}(\mathbf{y}) =
    \tilde{\mathbf{K}}_0^\intercal
            \mathbf{K}_0
            \mathbf{y}
    .
    \label{eq:chapter4_lwfsn_lin3}
\end{equation}
Assuming that $\mathbf{K}_0$ is a tight framelet i.e. $\tilde{ { \mathbf{K} } }_0^\intercal \mathbf{K}_0 = \mathbf{I} \cdot c$, with $c=1$, then
\begin{equation}
    \mathcalboondox{P} \{ L\}(\mathbf{y}) =
            \mathbf{y}
            .
    \label{eq:chapter4_lwfsn_lin4}
\end{equation}
This proves that the encoding-decoding section of the learned wavelet-frame shrinkage network allows for perfect signal reconstruction.
%
%%%%%%%%%%%%%%%%%%%%%%%%%%%%%%%%
\subsubsection{\textbf{Learned wavelet-frame shrinkage network -- number of operations}}
%%%%%%%%%%%%%%%%%%%%%%%%%%%%%%%%
%
The learned wavelet-frame shrinkage network contains a simpler convolution structure than the networks reviewed up to this moment. Therefore, for a single-level decomposition architecture, the total number of operations is
\begin{equation}
    \mathcal{O}(L) = 2 \cdot C_0 \cdot N_\text{r} \cdot N_\text{c} \cdot N_\text{f}^2~[\text{FLOPS}]
    .
\end{equation}
%
%%%%%%%%%%%%%%%%
\subsubsection{\textbf{Learned wavelet-frame shrinkage network -- residual variant}}
%%%%%%%%%%%%%%%%
%
\begin{figure}
    \centering
    \includegraphics[scale=0.75]{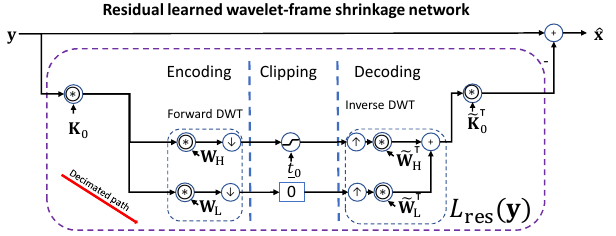}
    \caption{Residual version of the learned wavelet-frame shrinkage network. It can be noticed that the low-frequency branch of the network is nulled. In deeper networks it would further decomposed and the nulling would be activated at the deepest level (lowest resolution).}
    \label{fig:fig11}
\end{figure}
To illustrate the use of clipping activations in residual noise reduction, the residual version of the learned wavelet-frame shrinkage network is included. Note that there are two main differences with the conventional learned wavelet-frame shrinkage network. First, the shrinkage functions are replaced by clipping activations. Second, the low-frequency signal is suppressed. This is performed because the original design of the learned wavelet-frame shrinkage network does not have any non-linearities in that section. This is akin to the low-frequency nulling proposed by Kwon and Ye~\cite{kwon2021cycle}. The modified learned wavelet-frame shrinkage network is shown in Fig.~\ref{fig:fig11}. It can be observed that by setting to zero the low-frequency branch of the design the model is inherently assuming that the noise is high-pass.
%
%%%%%%%%%%%%%%%%%%%%%%%%%%%%%%%%
\subsubsection{\textbf{(Residual) Learned wavelet-frame shrinkage network -- concluding remarks}}
%%%%%%%%%%%%%%%%%%%%%%%%%%%%%%%%
The (residual) learned wavelet-frame shrinkage network is a design which explicitly mimics wavelet-shrinkage algorithms. It can be observed that the (r)LWFSN inherently assume that noise is high-frequency and explicitly avoid non-linear processing in the low-frequency band. Follow-up experiments also included non-linearities in the low-frequency band of the learned wavelet-frame shrinkage network~\cite{mondragon2022performance} and obtained similar results to the original design.
%
%
%
%
%%%%%%%%%%%%%%%%%%%%%%%%%%%%%%%%
\section{\textbf{What happens in trained models?}}\label{sec:trainedModel}
%%%%%%%%%%%%%%%%%%%%%%%%%%%%%%%%
%
\begin{figure}
    \centering
    \begin{minipage}{.5\columnwidth}
      \centering
      Initial $\mathbf{K}_2 \neq \tilde{\mathbf{K}}_2$
      \includegraphics[width=\columnwidth]{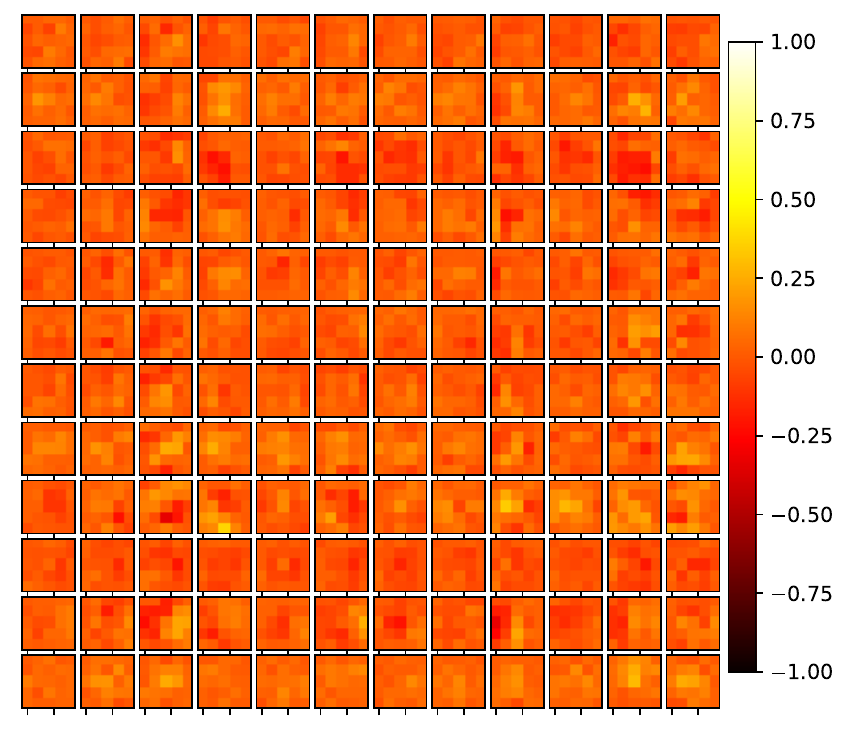}
    \end{minipage}%
    \begin{minipage}{.5\columnwidth}
        \centering
        Initial $\mathbf{K}_2 = \tilde{\mathbf{K}}_2$
        \includegraphics[width=\columnwidth]{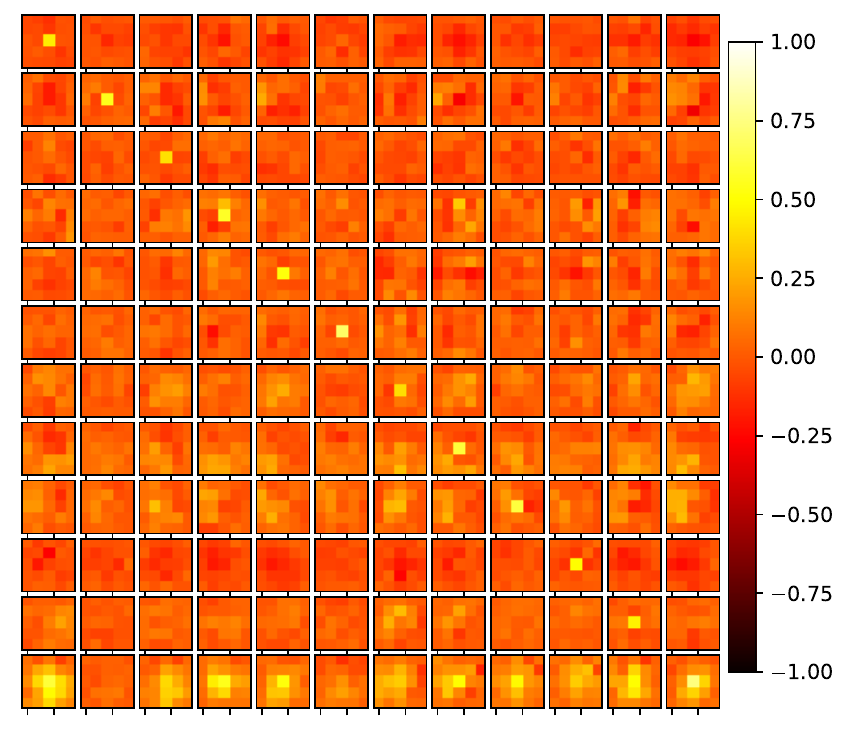}\\
    \end{minipage}
    \caption{Phase-complementary tight-framelet test for the trained toy network, initialized with random weights. The left figure shows the product $\tilde{\mathbf{K}}_2^\intercal(\mathbf{K}_2)_+$, where the initialization of $\mathbf{K}_2$ and $\tilde{\mathbf{K}}_2$ is different. It can be seen that the pair $(\mathbf{K}_2,\tilde{\mathbf{K}}_2)$ does not comply with the complementary-phase framelet criterion of Eq.~(\ref{eq:textbox_complementary}). This contrasts with the right result, which displays the result of the product $\tilde{\mathbf{K}}_2^\intercal(\mathbf{K}_2)_+$, for the same CNN, but where the initial values of $\tilde{\mathbf{K}}_2$ and $\mathbf{K}_2$ are identical. For this initialization, the filters approximate the complementary-phase tight-framelet criterion.}
    \label{fig:12}
\end{figure}
\begin{figure}
    \includegraphics[width=\columnwidth]{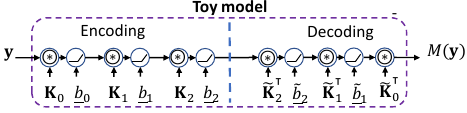}
    \caption{Toy model used for experiment on the properties of the filters of a trained CNN. The dimensions for tensors $\mathbf{K}_0$, $\mathbf{K}_1$ and $\mathbf{K}_2$ are $(6\times1\times3\times3)$, $(12\times6\times3\times3)$ and $(24\times12\times3\times3)$. The network is symmetric and the filter dimensions for the decoder convolution kernels $\tilde{\mathbf{K}}_n$ are the same as there corresponding encoding kernel $\mathbf{K}_n$.}
    \label{fig:fig13}
\end{figure}
%
%
%%%%%%%%%%%%%%%%%%%%%%%%%%%%%%%%
\subsection{\textbf{Properties of convolution kernels and low-rank approximation}}\label{sec:experiment1}
%%%%%%%%%%%%%%%%%%%%%%%%%%%%%%%%
%
The assumption that the convolution filters of a CNN behave as (complementary-phase) tight framelets is useful for analyzing the theoretical ability of a CNN to propagate signals. Albeit, it is difficult to prove that trained models comply with this assumption, because there are diverse elements affecting the optimization of the model, e.g. the initialization of the network, the  data presented to the model, the optimization algorithm as well as its parameters. In addition, in real CNNs, there may be co-adaptation between the diverse CNN layers, which may prevent that the individual filters of the CNN behave as tight framelets, since the decomposition and filtering performed by one layer is not independent from the rest~\cite{hinton2012improving}.

To test the behavior if the filters of trained CNN can converge to complementary-phase tight framelets, at least, on a simplified environment, we propose to train a toy model, as displayed in Fig.~\ref{fig:fig13}. If the trained filters of an encoder-decoder pair of the toy model ($\mathbf{K}_l$, $\tilde{\mathbf{K}}_l$), (where $l$ denotes one of the decomposition levels) behave as a complementary-phase tight framelet, then the pair ($\mathbf{K}_l$, $\tilde{\mathbf{K}}_l$) approximately complies with the condition presented in Eq.~(\ref{eq:frameletRedundancy}), which for identity input $\mathbf{I}$ simplifies to
\begin{equation}
    \tilde{\mathbf{K}}_n^\intercal(\mathbf{K}_n)_+  = \mathbf{I} \cdot c_n\,,
    \label{eq:textbox_complementary}
\end{equation}
in which $c_n$ is an arbitrary constant.

The toy model is trained on images that contain multiple randomly-generated overlapping triangles. All the images were scaled to the range [0,1]. For this experiment, the input to the images is the noise-contaminated image and the objective/desired output is the noiseless image. For training the CNNs, normally-distributed noise with standard deviation of 0.1 was added to the ground truth. For every epoch, a batch of 192~training images is generated. As validation and test images we use the ``astronaut" and ``cameraman" images included in the software package \emph{Scipy}. The model is optimized with Adam for 25~epochs with a linearly decreasing learning rate. The initial learning rate for the optimizer is set to 10$^{-3}$ and the batch size is set to 1~sample. The convolution kernels were initialized with Xavier initialization using a uniform distribution (see Glorot and Bengio~\cite{glorot2010understanding}). The code is available at IEEE's code ocean~\url{https://codeocean.com/capsule/7845737/tree}.
\begin{figure*}[t]
    \centering
    \includegraphics[scale=0.55]{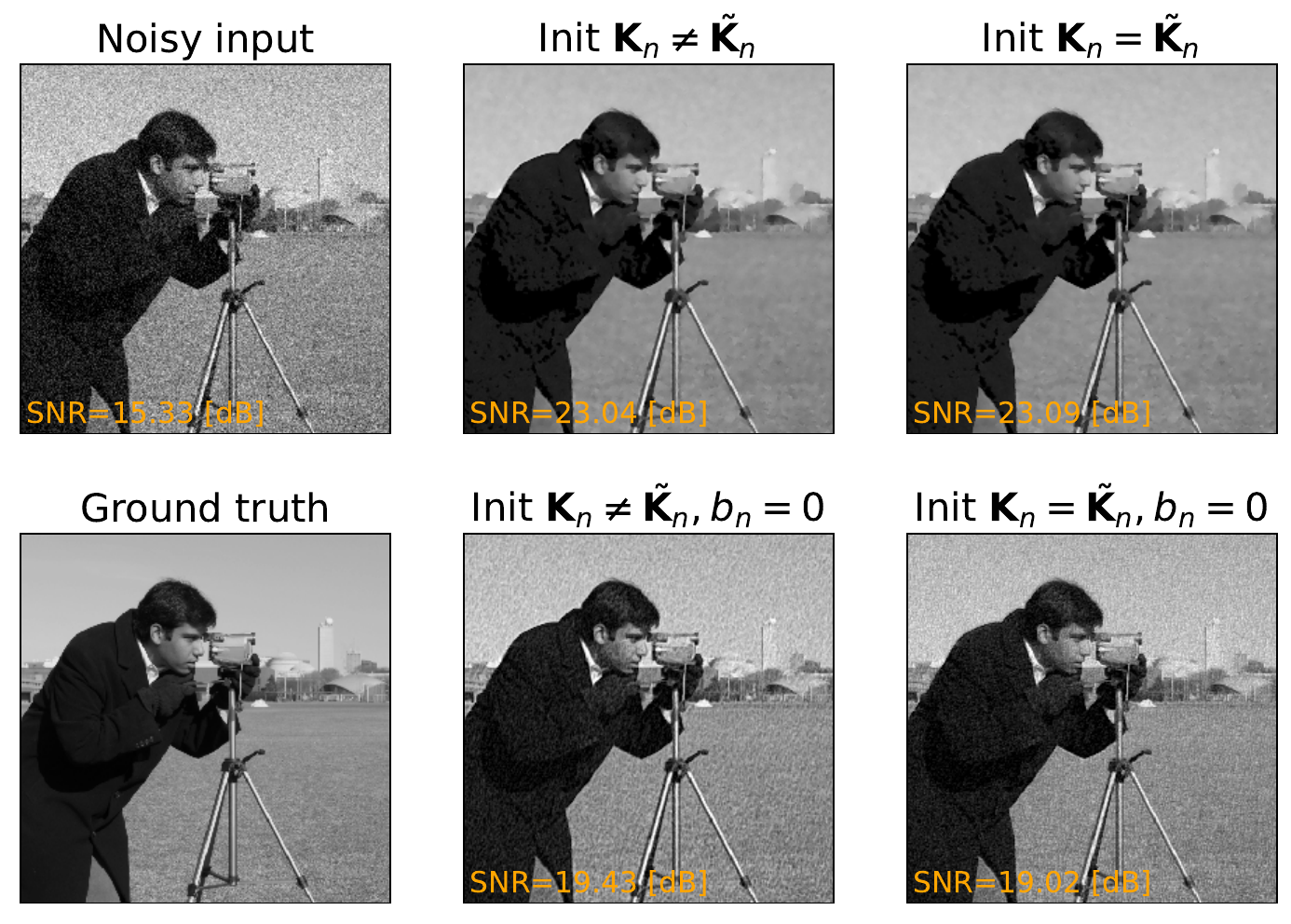}
    \caption{Processed ``cameraman" image for (in)dependently sampled initialization for the encoding and decoding filters. The top-left picture represents the noise-contaminated input ($\sigma_\eta=0.1$) and the bottom-left, the noiseless reference. The middle-column images are the processed noisy image with the toy model trained with different initialization for its convolution filters, while the right-column images are processed with the model where the same initial values are used for the encoding and decoding filters. The top-middle and top-right images are nearly identical in terms of quality and SNR, so that initialization has no effect. The middle and bottom-right images are the same model presented that processed the middle and top-right figures, but where its bias is set to zero. As expected, the noise is partly reconstructed.}
    \label{fig:fig14}
\end{figure*}

With the described settings, we have trained the toy model, and have tested if the phase-complementary tight-framelet property holds for the filters of the deepest level $l$=2. The results for the operation $\tilde{\mathbf{K}}_2^\intercal(\mathbf{K}_2)_+$ are displayed in Fig.~\ref{fig:12} (left), which shows that when the weights of the encoder and decoder have different initial values, the kernel pair ($\mathbf{K}_2$, $\tilde{\mathbf{K}}_2$) are not complementary-phase tight framelets. We have observed that the forward and inverse filters of wavelets/framelets are often the same or at least very similar. Based on this reasoning, we have initialized the toy model with the same initial values of the kernel pairs ($\mathbf{K}_n$, $\tilde{\mathbf{K}}_n$). As shown by Fig.~\ref{fig:12} (right), with the proposed initialization, the filters of the CNN  converge to tensors with properties reminiscent of complementary-phase tight-framelets. This suggests that the initialization of the CNN has an important influence on the convergence of the model to a specific solution.

Fig.~\ref{fig:fig14} displays a test image processed with two toy models, one trained trained with different and one trained with the same initial values for the encoding-decoding pairs. It can be observed that there are no significant differences between the images produced by both models. In the same figure (lower row), we have set the bias of both networks to zero. In this case, it is expected that the networks reconstruct the noisy input, as confirmed by the figure, where both CNNs partly reconstruct the original noisy signal. This result suggests that the ReLU plus bias pairs operate akin to the low-rank approximation mechanism proposed the theory of deep convolutional framelets.

The following conclusions can be drawn from this experiment. First, the filters of the CNN may not necessarily converge to complementary-phase tight framelets. This is possibly caused by the initialization of the network and/or the interaction/co-adaptation between the multiple encoder/decoder layers. Second, we confirm that for our experimental setting, the low-rank approximation behavior in the CNN can be observed. For example, when setting the biases and thresholds to zero, part of the noise texture (high-rank information) is recovered. Third, it is possible that linear filtering happens in the network as well, which may explain why noise texture is not fully recovered when setting the biases to zero. Fourth and final, we have observed that the behavior of the trained models change drastically depending on factors such as the learning rate and the initialization values of the model. For this reason, we consider this experiment and its outcome more as a proof of concept, where further investigation is needed. 
\begin{figure*}
    \centering
    \includegraphics[width=\textwidth]{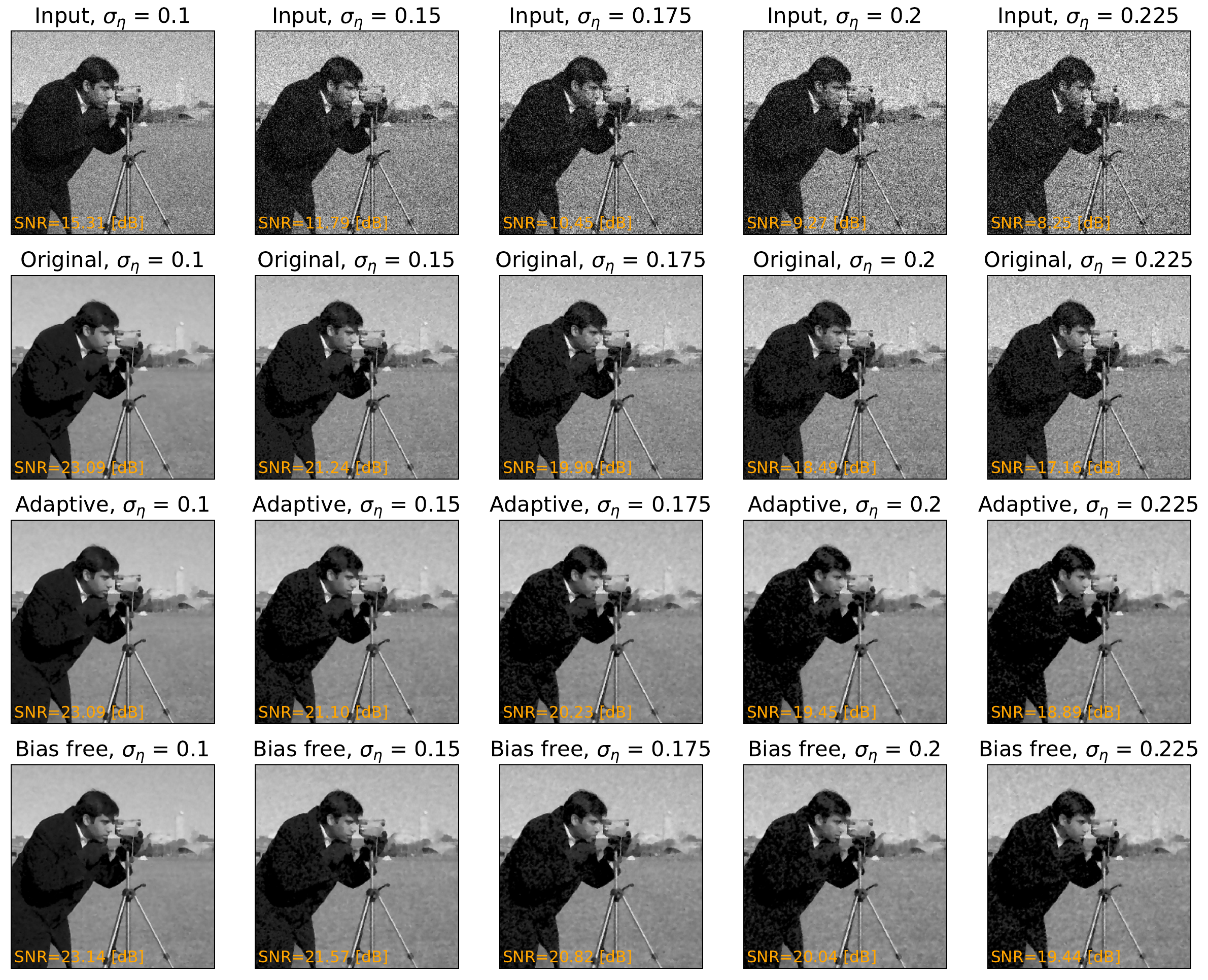}
    \caption{Comparison of the baseline (original) toy model against its adaptive and bias-free variants. The models are evaluated in the cameraman picture with increasing noise levels. The top row displays the noisy input. The second top-row represents the images processed with the original toy model. Meanwhile, the third row are the results of the adaptive toy model. Finally, the bottom-row are the results corresponding to the bias-free model. It can be observed that the performance original toy model degrades as the noise level increases, while the performance adaptive and bias-free model degrade less with increased noise levels, resulting in pictures with lower noise levels.}
    \label{fig:fig15}
\end{figure*}
%
%%%%%%%%%%%%%%%%%%%%%%%%%%%%%%%%
\subsection{\textbf{Generalization}}
%%%%%%%%%%%%%%%%%%%%%%%%%%%%%%%%
%
From the explanations in Section~\ref{sec:chapter4thresh}, it can be noted that the bias/threshold used in CNNs can modulate how much of the signal is suppressed by the nonlinearities. In addition, Section~\ref{sec:wiener} established that there are additional mechanisms for noise reduction within the CNN, such as the Wiener-like behavior observed by Mohan~\emph{et al.}~\cite{Mohan2020Robust}. This raises the question how robust conventional CNNs are to noise-level changes different from the level that the model has been trained with. To perform such experiment, we have trained two variants of the toy model. The first variant is inspired by the multi-scale sparse coding network by Mentl~\emph{et al.}~\cite{mentl2017noise}, where the biases of each of the nonlinearities (ReLU in this case) are multiplied by an estimate of the standard deviation of the noise. In the design of this example, the noise estimate $\hat{\sigma}_\eta$, which in accordance to Chang~\emph{et al.}~\cite{chang2000adaptive} is defined by
\begin{equation}
    \hat{\sigma}_\eta = 1.4826\cdot\text{Median}( \vert \mathbf{f}_\text{HH}*\mathbf{x} \vert ).
\end{equation}
Here, variable $\mathbf{f}_\text{HH}$ is the diagonal convolution filter of the discrete wavelet transform with Haar basis. For comparison purposes, we will refer to this model as \emph{adaptive} toy model. The second variant of the toy model being tested, examines the case where the convolution layers of the model do not add bias to the signal. This model is based in the so-called bias-free CNNs proposed by Mohan~\emph{et al.}, in which the bias of every convolution filter is set to zero during training. This setting  has the purpose of achieving better generalization on the model, since it is claimed that this modification causes the model to behave independently of the noise level.

We have trained the described variants of the toy models with the same settings of the experiment in Section~\ref{sec:experiment1}. The three models are evaluated on the test image with varying noise levels $\sigma_n\in[0.100, 0.150, 0.175, 0.200, 0.225]$, the result for this evaluation is displayed in Fig.~\ref{fig:fig15}. These results confirm that the performance of the original toy model degrades for higher noise levels. In contrast, the adaptive and bias-free toy model perform better than the original toy model for most noise levels.

The results of this experiment confirm the diverse noise-reduction mechanisms within a CNN, as well as showing that the CNNs have certain modeling limitations. For example, noise invariance, which can be addressed by further incorporating prior knowledge to the model, such as the case of the adaptive model, or by forcing the model to have a more Wiener-like behavior such as the case of the bias-free model. In the case of the bias-free model, note that, theoretically, it should be possible to obtain exactly the same behavior with the original toy model if the biases of the model would have converged to zero. This reasoning suggests that the large amount of free parameters and non-linear behavior of the model can potentially prevent to find the optimal/robust solution, in which case the incorporation of prior knowledge can help to improve the model.
%
%
%
%
%%%%%%%%%%%%%%%%%%%%%%%%%%%%%%%%
\section{ \textbf{Which network fits to  my problem?} }\label{sec:designElements}
%%%%%%%%%%%%%%%%%%%%%%%%%%%%%%%%
%
%
\begin{table*}[]
    \center
    \caption{Design elements and their impact in performance and computation cost.}
    \label{tab:designParameters}
    \begin{tabular}{ c c |c c c c  } 
    \toprule
    \multicolumn{2}{c|}{Design elements}  & Expressivity & Performance & No. parameters & Receptive field per layer\\
    \midrule                
    \multirow{3}{*}{        
        Activation}
                    & ReLU          & High & Best & High   & N/A\\ 
                    & Shrinkage     & Low  & Good & Medium & N/A\\
                    & Clipping      & Low  & Good & Medium & N/A\\ 
    \midrule
    \multirow{2}{*}{
        Scale}      & Single-scale  & High & Good & High       & Big\\
                    & Multi-scale   & High & Good & Medium/high& Small\\
    \midrule
    \multirow{2}{*}{
        Topology}   & Non-residual  & High & Good & Higher & N/A\\
                    & Residual      & High & Best & Lower  & N/A\\
    \bottomrule
    \end{tabular}
\end{table*}
%
%
%%%%%%%%%%%%%%%%%%%%%%%%%%%%%%%%
\subsection{\textbf{Design elements}}
%%%%%%%%%%%%%%%%%%%%%%%%%%%%%%%%
%
When choosing or designing a CNN for a specific noise-reduction application, multiple choices and design elements should be considered. For example, the required performance, the memory required to train/deploy models, if certain signal preservation characteristics are required, the target execution time for the model, the characteristics of the images being processed, etc. Based on these requirements diverse design elements of CNNs can be more or less desirable, for example, the activation functions, the use of single/multi-resolution models, the need for skip connections, and so forth. This section briefly discusses such elements by focusing on the impact that such elements have in terms of performance and potential computational cost. A summary of the main conclusions of these elements is included in Table~\ref{tab:designParameters}.
%
%%%%%%%%%%%%%%%%%%%%%%%%%%%%%%%%
\subsubsection{\textbf{Nonlinearity}}
%%%%%%%%%%%%%%%%%%%%%%%%%%%%%%%%
%
In literature, the most common activation function in CNNs is the ReLU. There are two main advantages of the ReLU with respect to other activations. First, ReLUs potentially enforce more sparsity in the feature maps than --for example-- the soft shrinkage, because ReLUs cancel not only small values of the feature maps like the shrinkage functions do, but also all the negative values. The second advantage of the ReLU is its capacity to approximate other functions (see Section~\ref{sec:shirnkageAndClippingReLU}). Note that the high-capacity of the ReLU to represent other functions~\cite{daubechies2022nonlinear, ye2019understanding} (often referred to as \emph{expressivity}) may also be one of the reasons why these models are prone to overfitting.

The better expressivity of the ReLU-CNNs may be the reason why --at the time of writing this manuscript-- ReLU-based CNNs perform marginally better than the shrinkage-based models in terms of metrics such as signal-to-noise ratio or the structural similarity index metric~\cite{scetbon2021deep, herbreteau2022dct2net, zavala2022noise}. Despite this small benefit, the visual characteristics of estimates produced by ReLU and shrinkage-based networks are very similar. Furthermore, the computational cost of ReLU-based designs is potentially higher those with shrinkage functions, because ReLUs require more feature maps to preserve the signal integrity. For example, the LWFSN shown in Section~\ref{sec:LWFSN} achieves a performance very close to the FBPConvNet and the tight-frame U-Net for noise reduction in computed tomography, but only with a small fraction of the total trainable parameters, which allows for a faster and less computation-expensive model~\cite{zavala2022noise}.

As concluding remark, it can be noted that regardless of the expressivity of the ReLU activation, it is not entirely clear if this means that ReLU activations outperform other functions such as the soft threshold \emph{in general}. Because we could not find articles, which are specifically focused on comparing the performance of ReLU/shrinkage-based models. In spite of this, there are some works that compare shrinkage-based CNNs with other (architecturally different) models based on ReLUs that indicate that the compared ReLU-based designs slightly outperform the shrinkage-based ones. For example, Herbeteau and Kevrann~\cite{herbreteau2022dct2net} proposed the so-called DCT2-Net, which is a shrinkage-based CNN, which despite of its good performance, it is still outperformed by the ReLU-based DnCNN~\cite{Zhang2017} CNN. A similar behavior was observed by Zavala~\emph{et al.}~\cite{zavala2022noise}, where their shrinkage-based LWFSN could not outperform the ReLU-based FBPConvNet~\cite{jin2017deep} and the tight-frame U-Net~\cite{han2018framing}. Another similar case is the deep K-SVD network~\cite{scetbon2021deep}, which achieves performance close (but slightly less good) than the ReLU-based DnCNN. Among the few examples were we found that a ReLU CNN performed better than shrinkage-based models, we find the work of Fan~\emph{et al.}~\cite{Fan2020}, where they compared variants of the so-called soft autoencoder and found that the shrinkage-based model outperformed the ReLU variant.
%
%%%%%%%%%%%%%%%%%%%%%%%%%%%%%%%%
\subsubsection{\textbf{Single/multi-scale designs}}
%%%%%%%%%%%%%%%%%%%%%%%%%%%%%%%%
%
Single-scale models have the advantage that they avoid aliasing because no down/up-sampling layers are used. Still this comes at the cost of more computations and memory. Furthermore, this approach may lead to models with larger filters and/or deeper networks to achieve the same receptive-field than multi-scale models, which may further increase the computation costs of single-scale models. 

In the case of multi-scale models, the main consideration should be that the down/up-sampling structure should allow perfect signal reconstruction to avoid introducing aliasing and/or distortion to the image estimates (e.g. the discrete wavelet transform in the tight-frame U-Net and in the learned wavelet-frame shrinkage network).
%
%%%%%%%%%%%%%%%%%%%%%%%%%%%%%%%%
\subsubsection{\textbf{(Non-) residual models}}
%%%%%%%%%%%%%%%%%%%%%%%%%%%%%%%%
%
Residual noise-reduction CNNs often perform better than their non-residual counterparts (e.g. the U-Net vs FBPConvNet and the LWFSN vs the rLWFSN). This may be caused because the trained models have more freedom to learn the filters, because the design does not need to learn to reconstruct the noiseless signal, but only to estimate the noise~\cite{ye2018deep}. Also, it can be observed that non-residual models potentially need more parameters than residual networks, because the propagation/reconstruction of the noiseless signal is also dependent on the number of channels of the network.
%
%%%%%%%%%%%%%%%%%%%%%%%%%%%%%%%%
\subsection{\textbf{State-of-the art}}
%%%%%%%%%%%%%%%%%%%%%%%%%%%%%%%%
%
Defining the state-of-the art in image denoising with CNNs is challenging for diverse reasons. First, there is a wide variety of available CNNs, which often are not compared to each other. Second, the suitability of a CNN for a given task may depend on the image and noise characteristics, such as noise distribution and (non-)~stationarity. Third, the large amount of variables, in terms of e.g. optimization, data and data augmentation, adds reproducibility issues which further complicate making a fair comparison between all available models~\cite{McCann2017convolutional}. In addition, it should be noted that for many of the existing models, the performance gap between state-of-the art models and other CNNs is often small.

Despite of the mentioned challenges, we have found some models that could be regarded as the state-of-the art. The first model to be addressed is the so-called DRU-Net~\cite{zhang2021plug}, which is a bias-free model~\cite{Mohan2020Robust} that incorporates a U-Net architecture with residual blocks. In addition, the DRU-Net uses an additional input to indicate to the network the noise intensity, which increases its generalization to different noise levels. An additional state-of-the art model is DnCNN~\cite{Zhang2017}. This network is residual, single-scale, while also using ReLU activations. Another state-of-the art model is the multi-level-wavelet CNN~\cite{liu2018multi}, which has a design very similar to the tight-frame U-Net~\cite{han2018framing}. Both of these models are based on the original U-Net design~\cite{ronnebergerUnet}, but are deployed in residual configuration and the down/up-sampling structure is based on the discrete wavelet transform. Furthermore, in addition to use encoding-decoding CNNs stand-alone, have used CNNs as proximal operator within model-based methods~\cite{monga2021algorithm, zhang2021plug}, which further improves the denoising power of non-model-based encoding-decoding CNNs.
%
%
%
%
%
%%%%%%%%%%%%%%%%%%%%%%%%%%%%%%%%
\section{\textbf{Conclusions and future outlook}}~\label{sec:chapter4_conclussions}
%%%%%%%%%%%%%%%%%%%%%%%%%%%%%%%%
%
In this paper, the widely used encoding-decoding CNN architecture has been analyzed from several signal processing principles. This analysis has revealed the following conclusions. (1) Multiple signal processing concepts converge in the mathematical formulation encoding-decoding CNNs models. For example, the convolution and down/up-sampling structure of the encoder-decoder structure is akin to the framelet decomposition, the activation functions are rooted on classical signal estimators. In addition, linear filtering also may happen within the model. (2) The activations implicitly assume noise and signal characteristics of the feature maps. (3) There are still many signal processing developments that can be integrated with current CNNs, further improving their performance in terms of accuracy, efficiency or robustness.

Despite of the signal processing nature of the encoding-decoding CNNs, at the moment of this publication, the integration of CNNs and existing signal processing algorithms is at an early stage. A clear example of the signal modeling limitations of current CNN denoisers are the activation functions, where the estimators provided by current activation layers neglect spatial correlation of the feature maps. Possible alternatives to solve this limitation could be to perform an activation function inspired on denoisers working on principles such as Markov random fields~\cite{malfait1997wavelet}, the locally-spatial indicators~\cite{Pizurica2006estimating} and multi-scale shrinkage~\cite{sendur2002bivariate}. Further ideas are provided by the extensive survey on denoising algorithms by Pi{\v{z}}urika and Philips~\cite{pivzurica1999image}. Additional approaches that can be further explored are non-local~\cite{dabov2008image} and collaborative filtering~\cite{ buades2005non}. Both techniques exploit the redundancy in natural images and only a few models are exploring these properties~\cite{yang2018bm3dnet, lee2022knn}.

To finalize this manuscript we encourage the reader to actively consider the properties of the signals processed, the design requirements and the existing signal processing algorithms when designing new CNNs. By doing so, we expect that next the generation of CNNs denoisers not only will be better performing, but also more interpretable, and reliable.
%
%
%
%
%%%%%%%%%%%%%%%%%%%%%%%%%%%%%%%%
\section*{\textbf{Acknowledgement}}
%%%%%%%%%%%%%%%%%%%%%%%%%%%%%%%%
We thank Dr.~Ulugbek Kamilov and the anonymous reviewers for their valuable feedback and suggestions for this article. 
%
%
%
%
%%%%%%%%%%%%%%%%%%%%%%%%%%%%%%%%
\bibliographystyle{IEEEtran}
%%%%%%%%%%%%%%%%%%%%%%%%%%%%%%%%
\bibliography{bibliography.bib}
\vskip -2.0\baselineskip plus -1fil
\begin{IEEEbiography}[{\includegraphics[width=1in,height=1.25in,clip,keepaspectratio]{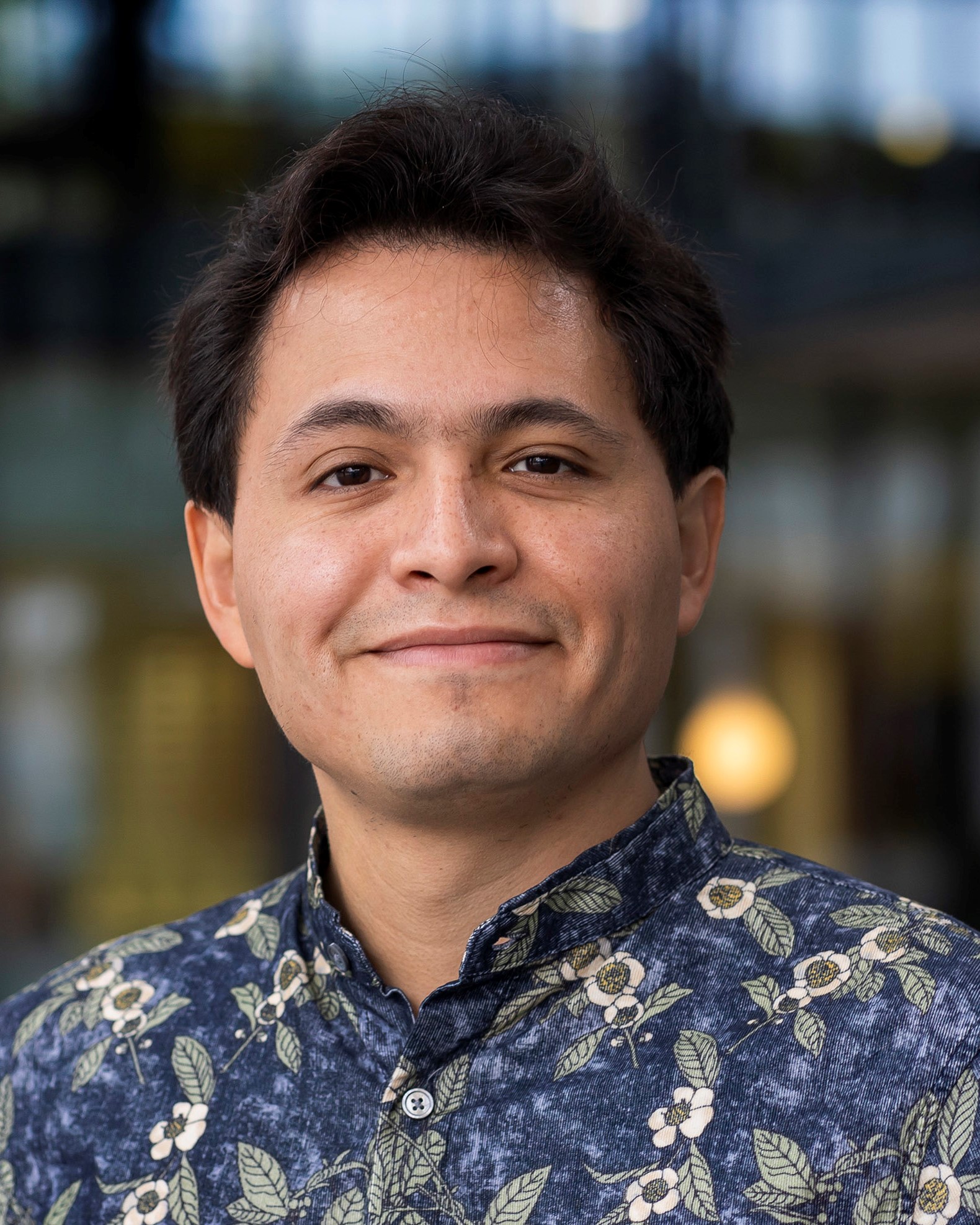}}]{Luis Albert Zavala-Mondragón}  (lzavala905@gmail.com)
is a PhD candidate at the Eindhoven University of Technology (TU/e). He holds a BSc and MSc in electrical engineering by the Universidad Nacional Autónoma de México (UNAM) and the Eindhoven University of Technology, respectively. Furthermore, he has experience in the field of hardware emulation (Intel, Mexico) and computer vision (Thirona, Netherlands). His current research interests are the development of efficient and explainable computer vision pipelines for healthcare applications. He is a  student member of IEEE.
\end{IEEEbiography}
\vskip -2.25\baselineskip plus -1fil
\begin{IEEEbiography}[{\includegraphics[width=1in,height=1.25in,clip,keepaspectratio]{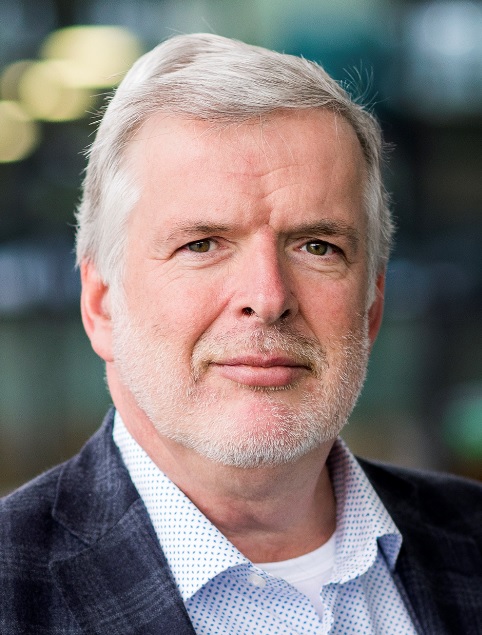}}]{Peter H. N. de With} (p.h.n.de.with@tue.nl) holds a MSc and PhD degree in electrical engineering from the Eindhoven University of Technology (TU/e) the Delft University of Technology, respectively. In 1984, he joined Philips Research Labs Eindhoven. In 1997-2000, he was a Full professor at University of Mannheim. From 2008 to 2010, he was VP of video technology at CycloMedia Technology.

In 2000 he became part-time professor at TU/e and since 2011, he is a full professor. He leads the Video Coding and Architectures (VCA) group. Peter has coauthored more than 70 refereed international book chapters and journal articles, over 500 conference publications and 40 international patents. He served as Technical Committee Member of the IEEE CES, ICIP, SPIE and is and co-recipient of multiple paper awards like the IEEE CES Paper Awards (several), VCIP and ICCE Best Paper Awards. He is member of the Royal Holland Society of Sciences and Humanities. He is a Fellow of IEEE.
\end{IEEEbiography}
\vskip -2.25\baselineskip plus -1fil
\begin{IEEEbiography}[{\includegraphics[width=1in,height=1.25in,clip,keepaspectratio]{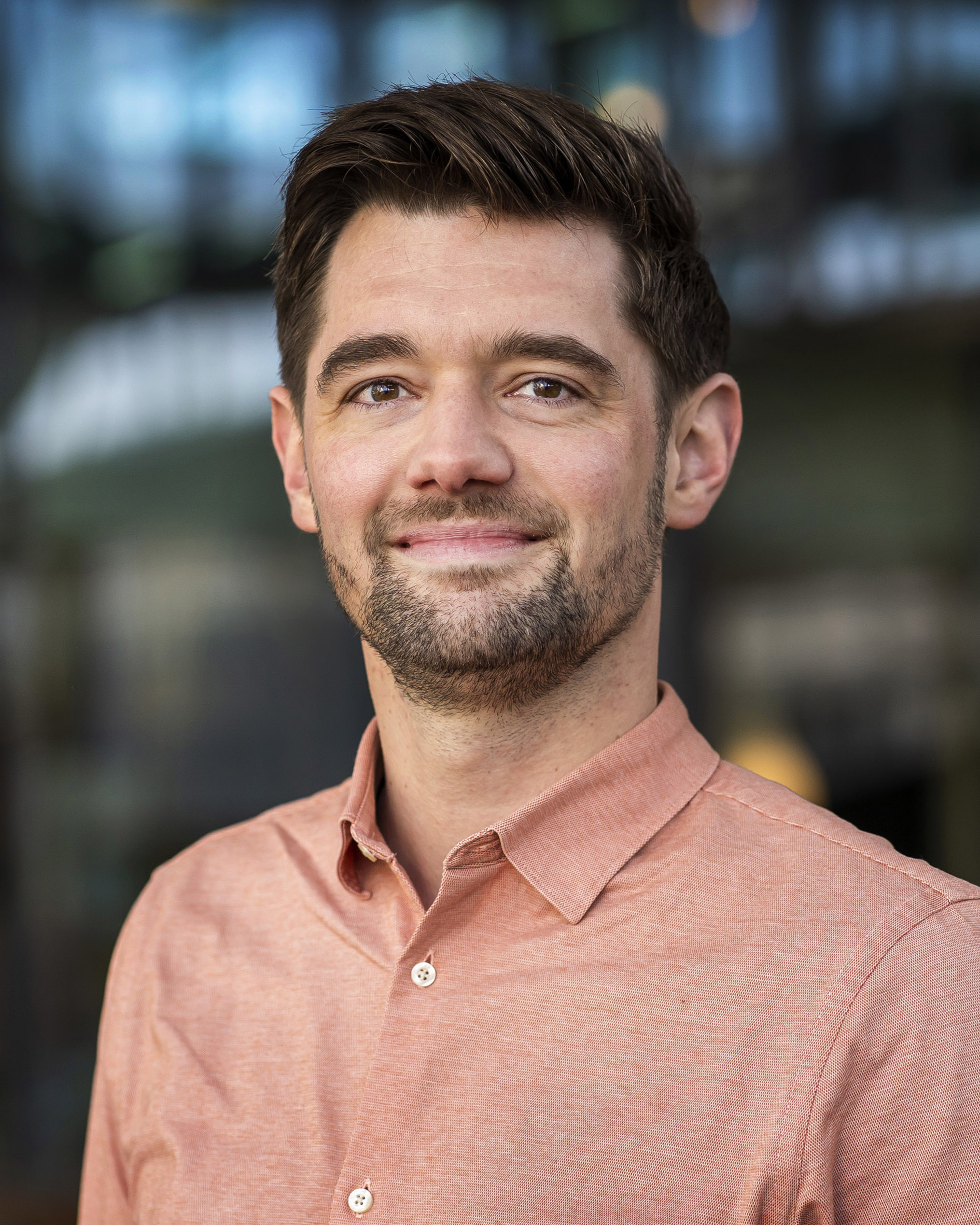}}]{Fons van der Sommen} (fvdsommen@tue.nl)
is an associate professor at Eindhoven University of Technology, holding a BSc and MSc in electrical engineering and a PhD in computer vision. Heading the healthcare cluster of the VCA research group, Fons has worked on a variety of image processing and computer vision applications, mainly in the medical domain. He has a strong interest in signal processing and information theory and strives to exploit methods from these fields to improve the robustness, efficiency and interpretability of modern-day AI architectures, such as Convolutional Neural Networks. He is a member of IEEE.
\end{IEEEbiography}
\vfill
\end{document}